\documentclass[9pt,twocolumn]{extarticle}

\usepackage[
  a4paper,
  top=2.0cm, bottom=2.0cm,
  left=1.5cm, right=1.5cm,
  columnsep=0.6cm
]{geometry}

\usepackage[T1]{fontenc}
\usepackage[utf8]{inputenc}
\usepackage{mathptmx}
\usepackage[scaled=0.91]{helvet}

\usepackage{microtype}

\usepackage{amsmath}
\usepackage{graphicx}
\usepackage[table]{xcolor}
\usepackage[colorlinks=true,linkcolor=black,citecolor=blue!60!black,urlcolor=blue!60!black]{hyperref}
\usepackage{longtable}
\usepackage{array}
\usepackage{booktabs}
\usepackage{multirow}
\usepackage{arydshln}
\usepackage{subcaption}
\usepackage{caption}
\usepackage{titlesec}
\usepackage{enumitem}
\usepackage{float}
\usepackage{setspace}
\usepackage{fancyhdr}
\usepackage{lastpage}
\usepackage{tikz}
\usepackage{etoolbox}
\usepackage{lineno}

\definecolor{NatCancerGray}{HTML}{5D6D7E}
\definecolor{RuleGray}{HTML}{B0B0B0}
\definecolor{SectionBlue}{HTML}{1B4F72}
\definecolor{grouprow}{gray}{0.92}

\DeclareCaptionType{listing}
\usetikzlibrary{arrows.meta, positioning, shapes, fit}

\titleformat{\section}
  {\large\bfseries\sffamily\color{SectionBlue}}
  {}
  {0em}
  {}
  [\vspace{1pt}{\color{RuleGray}\titlerule[0.5pt]}]

\titleformat{\subsection}
  {\normalsize\bfseries\sffamily\color{SectionBlue}}
  {}
  {0em}
  {}

\titleformat{\subsubsection}
  {\small\bfseries\itshape\sffamily\color{NatCancerGray}}
  {}
  {0em}
  {}

\titlespacing*{\section}      {0pt}{14pt plus 2pt minus 2pt}{5pt plus 1pt}
\titlespacing*{\subsection}   {0pt}{10pt plus 2pt minus 1pt}{3pt plus 1pt}
\titlespacing*{\subsubsection}{0pt}{6pt plus 1pt minus 1pt}{2pt}

\setlist{nosep,leftmargin=1.2em}

\fancypagestyle{firstpage}{%
  \fancyhf{}%
  \fancyfoot[C]{\small\sffamily\color{NatCancerGray}\thepage}%
}

\date{}

\newcommand{\maketitleblock}{%
\twocolumn[%
  \begin{flushleft}
  {\LARGE\bfseries\sffamily\color{black}%
  Routine laboratory trajectories encode the onset of organ-level complications in cancer\par}
  \end{flushleft}
  \vspace{6pt}
  \begin{flushleft}
  
  {\small\sffamily
  Jannik L\"ubberstedt\textsuperscript{1},
  Krischan Braitsch\textsuperscript{2},
  Jacqueline Lammert\textsuperscript{3,4},
  Christof Winter\textsuperscript{5},
  Florian Gabriel\textsuperscript{1},
  Tristan Lemke\textsuperscript{1},
  Christopher Zirn\textsuperscript{1},
  Markus Graf\textsuperscript{1},
  
  Friedrich Puttkammer\textsuperscript{1,6},
  Hartmut Häntze\textsuperscript{6},
  Johannes Moll\textsuperscript{1,7},
  Anirudh Narayanan\textsuperscript{1,6},
  Andrei Zhukov\textsuperscript{6},
  Fabian Drexel\textsuperscript{1},
  Zeineb Ben Chaaben\textsuperscript{1},

  Sebastian Ziegelmayer\textsuperscript{1},
  Su Hwan Kim\textsuperscript{1},
  Marion H\"ogner\textsuperscript{2},
  Jan Kirschke\textsuperscript{8},
  Florian Bassermann\textsuperscript{2,9,10,11},
  Marcus Makowski\textsuperscript{1},
  Christian Wachinger\textsuperscript{1},
  Lisa Adams\textsuperscript{1,*} \&
  Keno Bressem\textsuperscript{1,7,*}\par}
  \end{flushleft}
  \vspace{4pt}
  {\fontsize{7.5}{9.5}\selectfont\sffamily\color{NatCancerGray}%
  \textsuperscript{1}Department of Diagnostic and Interventional Radiology, School of Medicine and Health, TUM Klinikum, Rechts der Isar, Technical University of Munich, Germany.
  
  \textsuperscript{2}Department of Medicine III, School of Medicine and Health, TUM Klinikum, Rechts der Isar, Technical University of Munich, Germany.
  
  \textsuperscript{3}Chair of Medical Informatics, Institute of Artificial Intelligence in Medicine and Healthcare, School of Medicine and Health, TUM Klinikum, Rechts der Isar, Technical University of Munich, Germany.
  
  \textsuperscript{4}Clinical Department of Gynecology, School of Medicine and Health, TUM Klinikum, Rechts der Isar, Technical University of Munich, Munich, Germany.
  
  \textsuperscript{5}Department of Clinical Chemistry and Pathobiochemistry, School of Medicine and Health, TUM Klinikum, Rechts der Isar, Technical University of Munich, Germany.
  
  \textsuperscript{6}Department of Radiology, Charit\'e - Universit\"atsmedizin Berlin, Berlin, Germany.
  
  \textsuperscript{7}Department of Cardiovascular Radiology and Nuclear Medicine, School of Medicine and Health, TUM Klinikum, German Heart Center, Technical University of Munich. 

  \textsuperscript{8}Department of Neuroradiology, School of Medicine and Health, TUM Klinikum, Rechts der Isar, Technical University of Munich, Germany.
  
  \textsuperscript{9}TranslaTUM, Center for Translational Cancer Research, Technical University of Munich, Munich, Germany.
  
  \textsuperscript{10}Deutsches Konsortium für Translationale Krebsforschung, Heidelberg, Germany.
  
  \textsuperscript{11}Bavarian Cancer Research Center, Munich, Germany.

  \textsuperscript{*}Shared last authorship.\par}
  \vspace{10pt}
  \noindent{\color{RuleGray}\rule{\textwidth}{0.4pt}}
  \vskip 2pt
  \begin{minipage}{\textwidth}
  \small\sffamily
  \textbf{\color{SectionBlue}Abstract}\quad
Routine laboratory panels drawn during cancer treatment constitute longitudinal physiological recordings of organ function, yet their temporal structure is discarded by single-timepoint prognostic tools. A transformer trained on 2,777,595 laboratory measurements from 3,905 patients with multiple myeloma or ovarian cancer predicted the two-year onset of 162 treatment-associated complications, including therapy-related myelodysplastic syndromes, spanning eight clinical categories, achieving 1.5- to 6.1-fold enrichment above prevalence at the group level. It matched or outperformed non-sequential baselines across grouped endpoints (AUROC gains up to +0.11), demonstrating that longitudinal laboratory trajectories capture evolving complication-specific physiology inaccessible from isolated measurements. Predictions generalised across both cancers, divergence concentrating in disease-specific complications, and biomarker masking recovered signatures consistent with established pathophysiology. External validation on MIMIC-IV and MMRF CoMMpass confirmed transferability across independent healthcare systems (AUROC up to 0.85). Routine oncological laboratory data encode organ deterioration weeks to months before clinical onset, enabling complication-specific surveillance without additional testing infrastructure.

  \end{minipage}
  \vskip 2pt
  \noindent{\color{RuleGray}\rule{\textwidth}{0.4pt}}
  \vspace{10pt}
]%
}

\begin{document}
\thispagestyle{firstpage}
\maketitleblock

\section{Introduction}

Patients with advanced cancer who receive prolonged multi-line therapy progressively accumulate organ-level complications across multiple systems~\cite{Lustberg2023}, yet no existing model comprehensively predicts which specific complications a given patient will develop, or when~\cite{Hurria2011,Extermann2012}.
 The problem is most acute in diseases that require years of sequential therapy in predominantly older populations, where cumulative toxicity compounds pre-existing organ vulnerability. Multiple myeloma and ovarian cancer are prototypical examples, as both impose prolonged multi-line treatment on patients with median ages at diagnosis of approximately 69 years in multiple myeloma and 64 years in ovarian cancer~\cite{SEER_myeloma,Webb2017}, generating complication burdens that span renal dysfunction, marrow suppression, cardiovascular damage, recurrent infections, metabolic disease, and secondary malignancies~\cite{Terpos2018,Dimopoulos2023,Travis1999,Tuninetti2024,Malard2024}.

Throughout this treatment course, routine laboratory panels are drawn at nearly every clinical encounter. Complete blood counts, metabolic panels, liver and kidney function tests, coagulation parameters, and inflammatory markers generate dense longitudinal records that reflect the state of multiple organ systems simultaneously. These measurements are universally available, impose no additional testing burden, and capture physiological changes that may precede clinical manifestation of complications. Yet this temporal signal remains underexploited.  

The Cancer and Aging Research Group (CARG) score~\cite{Hurria2011,Hurria2016} and the Chemotherapy Risk Assessment Scale for High-Age Patients (CRASH)~\cite{Extermann2012} derive risk estimates from pre-treatment values; they neither update as physiology changes under therapy nor incorporate the trajectory of laboratory parameters across treatment lines, and they do not resolve individual complication types. External validation has reported discrimination as low as 0.52 (AUROC), in some populations~\cite{Frelaut2023,Moth2019}.

Disease-specific prognostic models, including recent transformer-based approaches for multiple myeloma that jointly predict survival and aggregate adverse events~\cite{hussain2024}, do not resolve the onset of individual treatment-associated complications, leaving clinicians without actionable guidance on which specific toxicities to anticipate in a given patient.
Generative transformers trained on large-scale electronic health records have demonstrated broad predictive capacity across general patient populations~\cite{Kraljevic2024,Shmatko2025,CoMET2025}, but represent laboratory values as discretized tokens and require richly-coded event streams that vary substantially in completeness across institutions. Laboratory-trajectory models in cancer populations predict only a small number of aggregate endpoints~\cite{Fu2025}. No existing model applies raw laboratory trajectories to complication-resolved risk estimation within defined cancer populations. A recent systematic review identified only seven studies that applied machine learning to longitudinal clinical data for cancer complication prediction~\cite{Zamani2025}, confirming that the temporal dimension of patient monitoring remains largely unexploited by current modeling approaches~\cite{Zhuang2025,swinckels2024}.

\begin{figure*}[ht!]
    \centering
    \includegraphics[width=\textwidth]{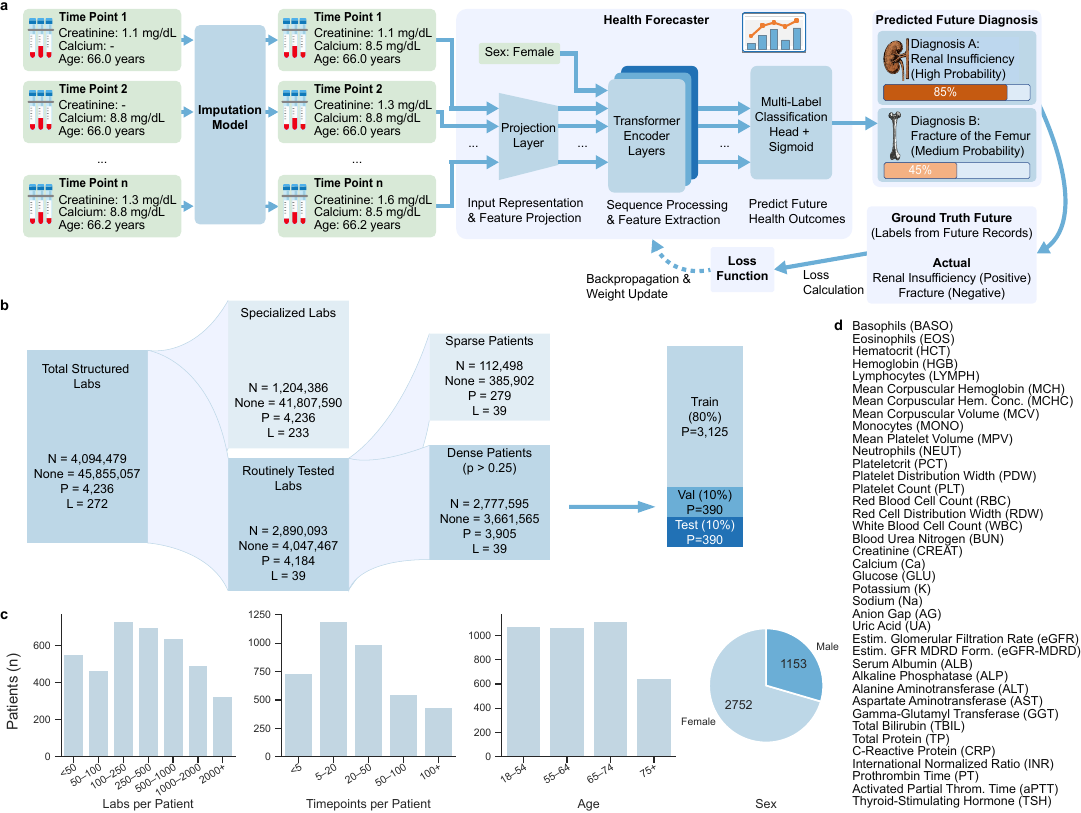}
    \caption{\textbf{Transformer-based health forecasting model for diagnosis prediction from longitudinal laboratory data.}
\textbf{a},~Architecture of the diagnosis prediction pipeline. Longitudinal laboratory values are aggregated into tokens representing randomly sampled time intervals of 10--30 days, each encoding z-score-normalised values of the 39 routinely tested analytes listed in \textbf{d} together with patient age. Missing values are completed by a dedicated imputation model. Token representations are projected into the model's embedding space via a projection layer and processed by a transformer encoder. A multi-label classification head with sigmoid activation predicts 162 future diagnoses within 2 years after the input interval finished, trained against ground-truth labels derived from subsequent clinical records.
\textbf{b},~Stepwise filtering of the laboratory dataset. Starting from all structured laboratory measurements ($N = 4{,}094{,}479$ values across $P = 4{,}236$ patients and $L = 272$ analytes), specialized laboratory tests were excluded to retain 39 routinely tested analytes ($N = 2{,}890{,}093$; $P = 4{,}184$ patients). Patients with sparse data coverage (proportion of observed values $\leq 0.25$) were removed, yielding a final dataset of $N = 2{,}777{,}595$ measurements across $P = 3{,}905$ patients, split into training (80\%, $P = 3{,}125$), validation (10\%, $P = 390$) and test (10\%, $P = 390$) sets.
\textbf{c},~Cohort characteristics of the final dataset, showing distributions of the number of individual laboratory measurements per patient, number of aggregated time points per patient, patient age and sex.
\textbf{d},~Complete list of the 39 routinely tested laboratory analytes used as model inputs, spanning haematology, metabolic, hepatic, inflammatory, coagulation and endocrine panel (Extended Data Table~\ref{tab:lab_inputs}).}
\label{fig:health_forecaster}
\end{figure*}
This study tested whether routine laboratory trajectories encode complication-specific temporal signal by training a transformer on 2,777,595 measurements from 3,905 patients with multiple myeloma or ovarian cancer, segmented into observation windows of 10–30 days, to predict the two-year onset of 162 treatment-associated complications. 
These two malignancies were chosen because they share the burden of cumulative treatment toxicity but diverge mechanistically in informative ways. Renal impairment is treatment-limiting in both yet arises from light-chain cast nephropathy in myeloma versus platinum nephrotoxicity in ovarian cancer~\cite{Dimopoulos2023,Perazella2012}, and infection risk carries a disease-intrinsic component in myeloma, through clonal suppression of immunoglobulin production, that is absent in ovarian cancer~\cite{Raje2022}. This combination of shared and disease-specific endpoints provided a testbed for whether laboratory-based prediction captures biologically coherent signal or relies on cohort-level confounders.
Predictive performance was evaluated across both malignancies, compared against non-sequential baselines, interrogated through biomarker masking to identify which laboratory analytes drive complication-specific predictions, and tested for transferability against two independent United States cohorts, the MIMIC-IV critical-care database and the MMRF CoMMpass multiple myeloma registry. If this temporal signal exists, the data infrastructure for its detection is already in place in standard oncological care.

\section{Results}

\begin{figure*}[ht!]
        \centering
        \includegraphics[width=\textwidth]{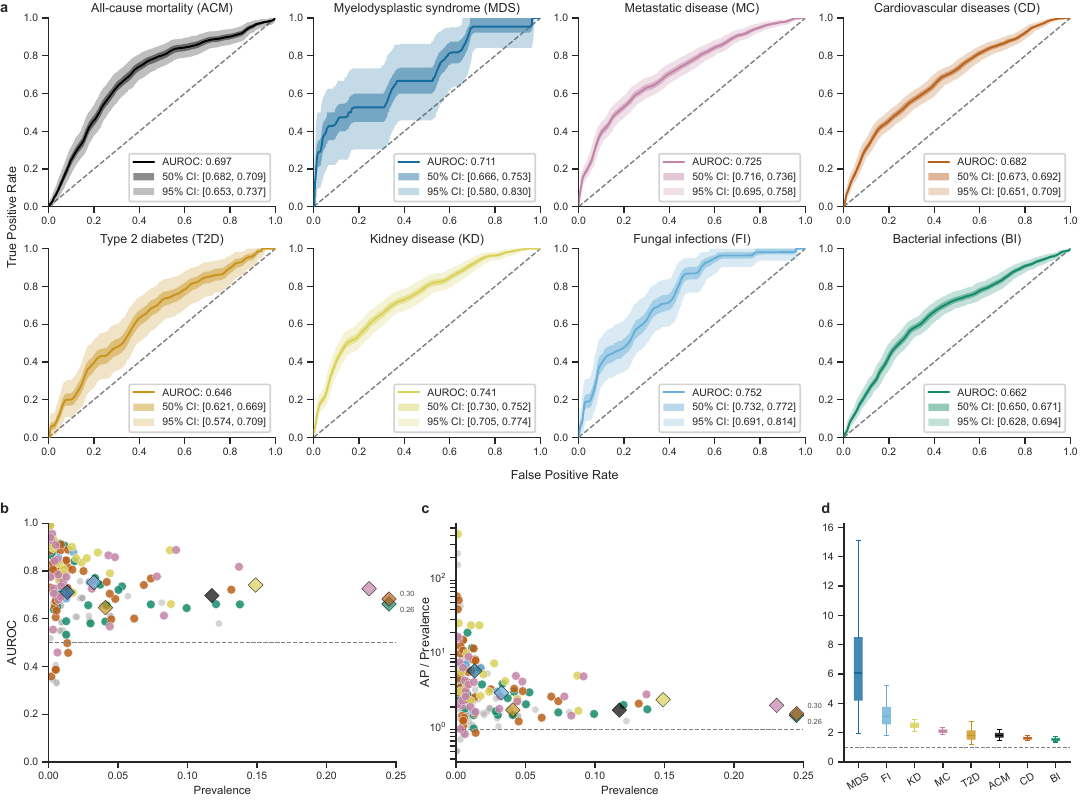}
\caption{\textbf{Diagnosis prediction performance across clinical endpoints and disease groups.}
\textbf{a},~Receiver operating characteristic curves for the prediction of eight clinically relevant diagnoses and disease groups within a 2-year horizon. Solid lines indicate median performance across bootstrap resamples. Shaded regions denote 50\% and 95\% bootstrap confidence intervals. The dashed diagonal represents random classifier performance (AUROC = 0.5).
\textbf{b},~AUROC as a function of prevalence for individual ICD-10 diagnosis codes. Each point represents a single diagnosis; diamonds indicate grouped endpoint performance. Colours indicate disease group membership, grey shows without memberships in the eight primary groups. The dashed horizontal line at 0.5 indicates random discrimination. The cardiovascular disease (CD; prevalence = 0.30) and bacterial infection groups (BI; prevalence = 0.26) exceed the x-axis range and are plotted at the right edge with their prevalence indicated. Model discrimination consistently exceeds chance for higher-prevalence diagnoses, where estimation variance is lower. For rare diagnoses, confidence intervals widen but the model still achieves above-chance discrimination.
\textbf{c},~Enrichment ratio (AP/prevalence, log$_{10}$ scale) as a function of prevalence for each individual diagnosis. The cardiovascular disease (CD; prevalence = 0.30) and bacterial infection groups (BI; prevalence = 0.26) exceed the x-axis range and are plotted at the right edge with their prevalence indicated. The model achieves the largest relative gains for rare diagnoses, with AP exceeding prevalence by up to an order of magnitude for the lowest-prevalence conditions. Diamonds indicate grouped endpoints.
\textbf{d},~Enrichment ratio (AP/prevalence, linear scale) with 50\% and 95\% bootstrap confidence intervals for each grouped diagnostic endpoint, ordered by decreasing enrichment. The dashed line at 1.0 indicates the prevalence baseline. Endpoints above this line are predicted at above-chance enrichment, with myelodysplastic syndrome (MDS) and fungal infections (FI) showing the strongest relative gains.}
\label{fig:diagnosis_prediction}
\end{figure*}
\subsection{Cohort characteristics}

Across 39 routinely measured analytes, 2,777,595 structured laboratory measurements were extracted from the hospital information system of a single tertiary centre, spanning 3,905 patients (1,855 multiple myeloma, 2,050 ovarian cancer) treated between 2000 and 2026. Patients with sparse coverage (<25\%) were excluded and only parameters from standard clinical assessment retained (Figure~\ref{fig:health_forecaster}b,c; Extended Data Table~\ref{tab:lab_inputs}), then split into non-overlapping training ($P=3{,}125$), validation ($P=390$), and test ($P=390$) sets. Clinical documents from the myeloma subcohort were separately processed by an LLM-based extraction pipeline for cohort characterisation only, not as model input (Supplementary Results, Extended Data Fig.~\ref{fig:structuring_overview}; comparison against existing myeloma cohorts in Extended Data Table~\ref{tab:mm_datasets_comparison}).
Laboratory trajectories were segmented into non-overlapping 10 to 30 day intervals. Patients contributed a median of 16 timepoints (IQR 7--26) over 1,091 days (IQR 360--2,682). Missing values were completed by a transformer-based imputation model (see Methods); across the final dataset, 43.1\% of analyte-timestep entries were observed. Diagnostic endpoints were curated from approximately 3,000 ICD-10 codes, retaining 162 diagnoses in eight categories: myelodysplastic syndrome, cardiovascular disease, kidney disease, fungal infections, bacterial infections, metastatic disease, type 2 diabetes, and all-cause mortality (Figure~\ref{fig:health_forecaster}a). Performance on a held-out test set of 390 patients (1,579 sequences) was evaluated using AUROC and enrichment ratio (AP divided by prevalence) (Supplementary Tables~\ref{tab:performance_comparison}--\ref{tab:performance_comparison_extern_grouped}; see Methods).

\begin{figure*}[ht!]
        \centering
        \includegraphics[width=\textwidth]{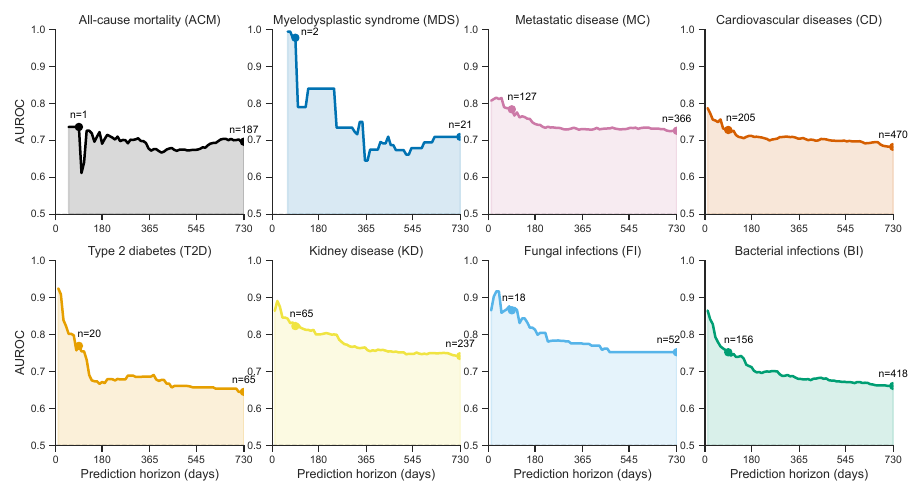}
\caption{\textbf{Cumulative diagnosis-prediction performance as a function of forecasting horizon.}
    Area under the receiver operating characteristic curve (AUROC) for the Transformer-based health forecaster as a function of the prediction horizon, defined as the number of days after the end of the laboratory-input window within which the diagnosis occurs. At each horizon $n$, performance is evaluated cumulatively: a case is counted as positive if the diagnosis is recorded at any point from the end of the input window up to and including day $n$. The maximum prediction horizon is 730 days. Each panel corresponds to one of the eight grouped diagnoses. Curves show AUROC, with the filled region denoting the area under each curve. Annotations indicate the number of positive cases at 90 and 730s days.}
    \label{fig:horizon_auroc}
\end{figure*}
\subsection{Prediction of treatment-associated complications}
The transformer predicted the two-year onset of all eight primary disease groups above their respective prevalence baselines (Figure~\ref{fig:diagnosis_prediction}). AUROC ranged from 0.65 for type 2 diabetes to 0.75 for fungal infections, and predictive enrichment was inversely related to prevalence, with the highest ratios observed for the rarest endpoints, reaching up to an order of magnitude for diagnoses with prevalence below 0.01 (Figure~\ref{fig:diagnosis_prediction}b,c,d). The strongest enrichment among grouped endpoints was observed for myelodysplastic syndromes (AUROC 0.71; 6.1-fold, CI 95\%=[1.9,15.1]), followed by fungal infections (AUROC 0.75; 3.1-fold), kidney disease (AUROC 0.74; 2.5-fold), and metastatic disease (AUROC 0.73; 2.1-fold). All-cause mortality (AUROC 0.70; 1.8-fold), type~2 diabetes (AUROC 0.65; 1.8-fold), cardiovascular diseases (AUROC 0.68; 1.6-fold), and bacterial infections (AUROC 0.66; 1.5-fold) showed lower enrichment (Supplementary Table~\ref{tab:performance_comparison_grouped}). Individual cardiovascular endpoints showed stronger enrichment and discrimination when evaluated separately, with chronic ischaemic heart disease reaching 2.5-fold enrichment (AUROC 0.71), atrial fibrillation 2.8-fold (AUROC 0.74), and heart failure 3.3-fold (AUROC 0.68).

\begin{figure*}[ht!]
    \centering
    \includegraphics[width=\textwidth]{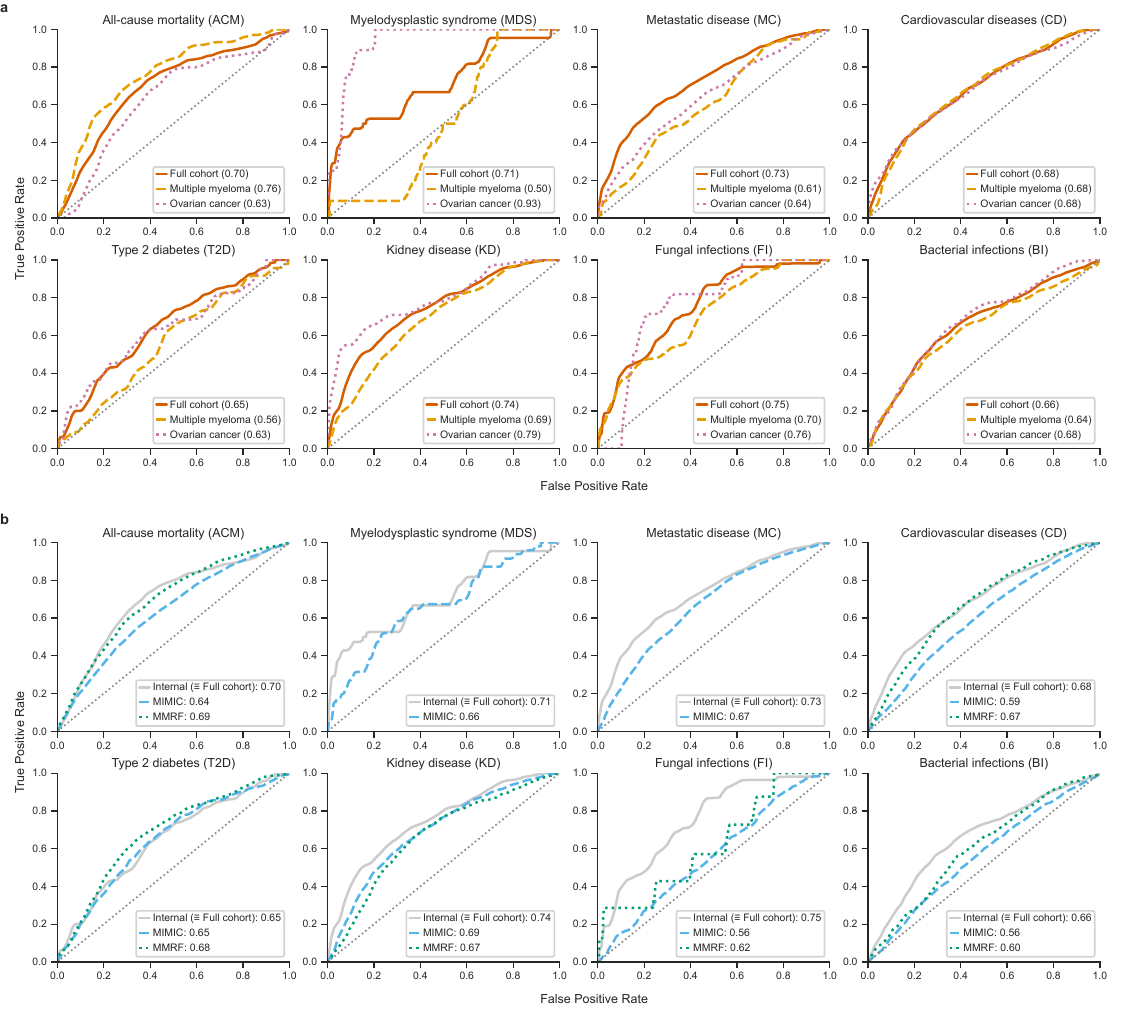}
\caption{\textbf{Diagnosis prediction performance across cancer-type subcohorts and Datasets.}
\textbf{a} Receiver operating characteristic (ROC) curves for the eight diagnosis groups evaluated on the full cohort (full cohort; $P = 390$ patients, 1{,}579 sequences), the multiple myeloma subcohort ($P = 182$, 853 sequences) and the ovarian carcinoma subcohort ($P = 208$, 726 sequences). Values in parentheses denote AUROC; the dashed diagonal represents random performance (AUROC 0.5). Five of eight endpoints generalised across subcohorts (AUROC differences $\leq$0.07); kidney disease, all-cause mortality, and myelodysplastic syndromes diverged. \textbf{b}
ROC curves for eight diagnosis groups evaluated
on the internal held-out test set ($N_{\mathrm{eval}} = 1{,}579$ sequences), MIMIC-IV
($N_{\mathrm{eval}} = 8{,}150$ sequences; v3.1) and MMRF CoMMpass ($N_{\mathrm{eval}} = 7{,}347$
sequences; IA24). Values in parentheses denote area under the
receiver operating characteristic curve (AUROC). The dashed diagonal
represents random classifier performance (AUROC = 0.5). Discrimination is
preserved across cohorts for all-cause mortality, type 2 diabetes and kidney
disease, and additionally on MMRF for cardiovascular diseases, and on MIMIC-IV for MDS. Selective
divergence on MIMIC-IV is observed for categories sensitive to coding-system
differences, most prominently cardiovascular, bacterial and fungal endpoints.
Myelodysplastic syndrome and metastatic disease are not shown for the MMRF
subcohort, as both are not provided diagnostic labels.}

\label{fig:subcohort_comparison}
\end{figure*}

\subsection{Predictive performance across pre-diagnostic time horizons}

To assess performance as a function of lead time, the model was evaluated cumulatively across forecasting horizons from 10 to 730 days (Figure~\ref{fig:horizon_auroc}). At each horizon, patients were labelled positive if the target diagnosis occurred between the end of the laboratory-input window and the specified prediction horizon.

Across the aggregate outcome, AUROC declined gradually with increasing forecasting horizon, from 0.682 at 10 days to 0.644 at 730 days. The grouped endpoints showed the same qualitative trend but differed in the rate of decline. Kidney disease, metastatic disease and fungal infections retained the highest long-horizon discrimination, remaining at AUROCs of 0.741, 0.725 and 0.752, respectively, at 730 days. Cardiovascular disease declined from 0.787 to 0.682 across the same interval, whereas bacterial infections and type~2 diabetes declined more steeply. All-cause mortality followed a distinct trajectory, becoming evaluable only after approximately 50 days owing to few early events, and stabilizing near 0.70 at longer horizons.

\begin{figure*}[ht!]
        \centering
        \includegraphics[width=\textwidth]{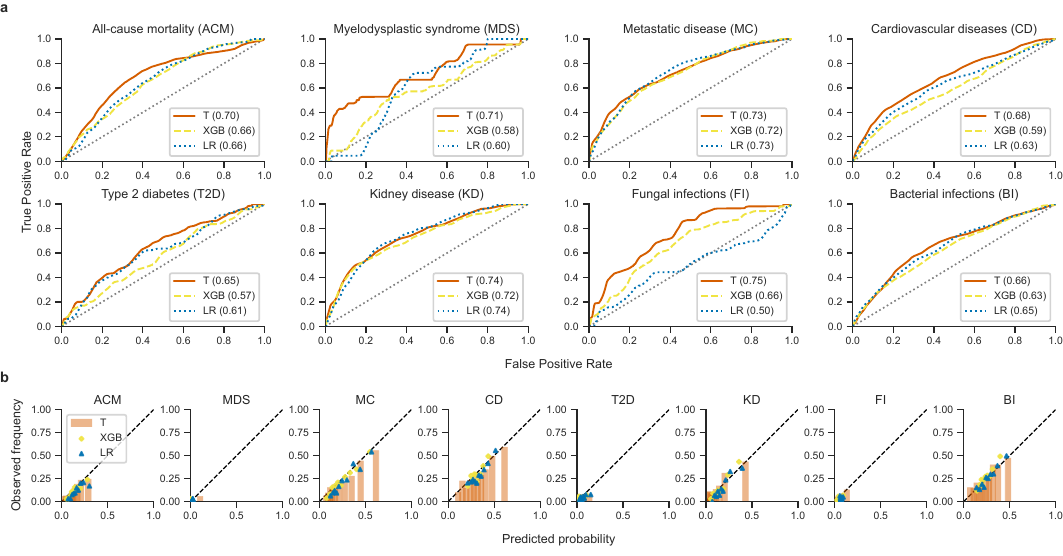}
\caption{\textbf{Model performance and calibration.}
\textbf{a},~Receiver operating characteristic curves for the transformer (T), XGBoost (XGB) and logistic regression (LR) baselines across eight diagnosis groups. Values in parentheses denote the area under the receiver operating characteristic curve (AUROC). The XGBoost and logistic regression architectures follow the implementations described in the \emph{CoMET} framework, where XGBoost performed comparably to CoMET-L for incident disease prediction in low-prevalence settings. The transformer matches or exceeds both baselines on all eight endpoints, with gains over the strongest baseline of up to $+$0.11 for myelodysplastic syndrome and $+$0.09 for fungal infections; for metastatic and kidney disease it performs on par with logistic regression (AUROC 0.73 and 0.74, respectively).
\textbf{b},~Per-endpoint calibration plots for all three models after Platt scaling on the validation set, fitted independently for each diagnosis group. Predictions are binned into ten equally sized quantiles. Each marker represents the mean predicted probability (x-axis) versus the observed event frequency (y-axis) within one quantile bin for the transformer (bars), XGBoost (diamonds) and logistic regression (triangles). The dashed diagonal indicates perfect calibration. All three models are well calibrated and overlap closely, tracking the diagonal across the observed range of predicted probabilities. For the rare endpoints (myelodysplastic syndrome, type 2 diabetes and fungal infections) predictions concentrate near the origin, consistent with low event prevalence, whereas for the more prevalent endpoints (all-cause mortality, metastatic disease, cardiovascular disease, kidney disease and bacterial infections) predictions span a wider range and remain close to the diagonal, with the highest-probability bins falling slightly below it.}
\label{fig:model_comparison}
\end{figure*}

\subsection{Biological coherence across malignancies}

To test whether predictions reflected biologically coherent signal rather than cohort-level confounders, the model was evaluated separately on the multiple myeloma (MM; P = 182 patients, 853 sequences) and ovarian carcinoma (OC; P = 208 patients, 726 sequences) test subsets (Figure~\ref{fig:subcohort_comparison}; Supplementary Table~\ref{tab:performance_comparison_grouped}).
Five of eight grouped endpoints generalised across both malignancies with comparable discrimination. Cardiovascular diseases (MM: AUROC 0.68; OC: AUROC 0.68), bacterial infections (MM: AUROC 0.64; OC: AUROC 0.68), metastatic disease (MM: AUROC 0.61; OC: AUROC 0.64), type~2 diabetes (MM: AUROC 0.56; OC: AUROC 0.63), and fungal infections (MM: AUROC 0.70; OC: AUROC 0.76) showed AUROC differences of no more than 0.07 across subcohorts despite differing prevalences. For metastatic disease, full-cohort discrimination (AUROC 0.73) exceeded both subcohort estimates, consistent with the model leveraging cohort-level prevalence differences in addition to within-cohort signal; C78 (secondary malignant neoplasms of respiratory and digestive organs) accounted for 28\% prevalence in the ovarian subcohort but only 1\% in myeloma, with correspondingly divergent average precision (OC: 0.40; MM: 0.02). Fungal infection discrimination was preserved across subcohorts despite a disease-asymmetric pathogen distribution, with pulmonary aspergillosis (B44.1) confined to the myeloma subcohort (MM: AUROC 0.84; OC: zero prevalence) and candidiasis (B37) captured in both populations (MM: AUROC 0.71; OC: AUROC 0.77).
The remaining three endpoints diverged. Kidney disease showed a moderate cross-subcohort gap (MM: AUROC 0.69; OC: AUROC 0.79), while all-cause mortality showed a larger gap in the opposite direction (MM: AUROC 0.76; OC: AUROC 0.63). Myelodysplastic syndromes separated most sharply (OC: AUROC 0.93, 20.2-fold enrichment; MM: AUROC 0.50, 1.3-fold enrichment), with limited case counts ($N=21$), precluding stable estimation.

\begin{figure*}[ht!]
        \centering
        \includegraphics[width=\textwidth]{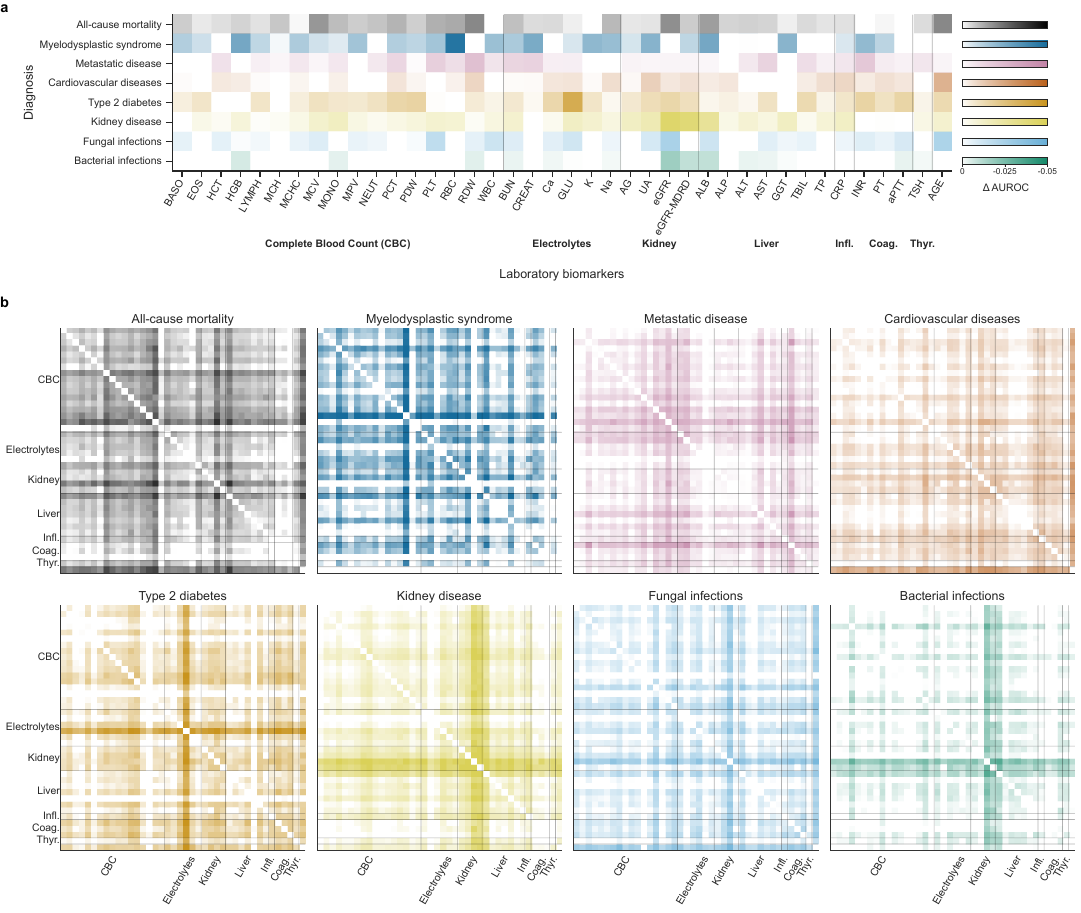}
\caption{\textbf{Feature importance analysis reveals diagnosis-specific biomarker contributions.}
\textbf{a},~Change in area under the receiver operating characteristic curve ($\Delta$AUROC) when individual laboratory biomarkers or age are masked, shown per diagnosis group. Colour intensity reflects the magnitude of discrimination loss upon removal. Stronger shading indicates greater importance of the corresponding feature for predicting the given diagnosis. Features are grouped by laboratory panel (complete blood count, electrolytes, kidney, liver, inflammatory, coagulation and thyroid markers).
\textbf{b},~Pairwise feature masking analysis. Each matrix shows the $\Delta$AUROC when pairs of input features are simultaneously removed, for each of the eight predicted diagnosis groups. Off-diagonal entries reveal synergistic or redundant relationships between biomarkers. Strongly coloured off-diagonal cells indicate feature pairs whose joint removal causes a disproportionate discrimination loss beyond their individual contributions. Diagnosis-specific patterns emerge, such as the pronounced importance of kidney function markers for kidney disease prediction and of haematological parameters for myelodysplastic syndrome.}
\label{fig:feature_importance}
\end{figure*}
\subsection{Temporal advantage over non-sequential baselines}
Two non-sequential baselines, XGBoost and logistic regression, were trained on tabular representations of the same laboratory sequences following the baseline evaluation protocol described in CoMET~\cite{CoMET2025} (see Methods). The transformer matched or exceeded both XGBoost and logistic regression for all eight grouped endpoints, outperforming both baselines on six of them (Figure~\ref{fig:model_comparison}a; Supplementary Tables~\ref{tab:performance_comparison} and ~\ref{tab:performance_comparison_grouped}).
The magnitude of the transformer's AUROC advantage over the best non-sequential baseline varied across endpoints. The largest gain was observed for myelodysplastic syndromes (transformer AUROC 0.71 vs.\ logistic regression 0.60; $\Delta$AUROC $+$0.11), followed by fungal infections (AUROC 0.75 vs.\ XGBoost 0.66; $+$0.09). Cardiovascular diseases (AUROC 0.68 vs.\ logistic regression 0.63; $+$0.05), all-cause mortality (AUROC 0.70 vs.\ XGBoost and logistic regression, both 0.66; $+$0.04), and type~2 diabetes (AUROC 0.65 vs.\ logistic regression 0.61; $+$0.04) showed intermediate gains. Bacterial infections showed the smallest gain (AUROC 0.66 vs.\ logistic regression 0.65; $+$0.01). For metastatic disease (AUROC 0.73) and kidney disease (AUROC 0.74), the transformer performed on par with logistic regression (0.73 and 0.74, respectively). Model calibration is reported in Supplementary Results.

\subsection{Biomarker signatures}
To identify which laboratory analytes drove complication-specific predictions, single-feature and pairwise masking analyses were performed on the test set, recording the change in AUROC ($\Delta$
AUROC) when an analyte (or pair) and its validity indicator were zeroed across all timesteps (Figure~\ref{fig:feature_importance}).

Single-feature masking produced diagnosis-specific patterns that varied in concentration. Three endpoints had a clearly dominant analyte: glucose for type~2 diabetes ($-$0.036), red blood cell count for myelodysplastic syndromes (
$-$0.046, the largest single drop in the matrix, ahead of the more conventional haemoglobin at $-$0.026), and the glomerular filtration estimates for kidney disease (eGFR $-$0.043, eGFR-MDRD $-$0.039). For the remaining endpoints, discrimination loss was distributed across tight clusters of contributors rather than a single dominant biomarker (all-cause mortality, cardiovascular, metastatic, and fungal infections), with bacterial infections showing the weakest single-feature signal, led by eGFR. Across endpoints, albumin was the most cross-cutting contributor (top five in five of eight), eGFR led three (kidney, fungal, bacterial), and age led two (cardiovascular, all-cause mortality).

Pairwise masking confirmed these patterns. The eGFR-MDRD/eGFR pair ($-$0.086) was the largest pairwise drop in the matrix, and all five top kidney pairs included a filtration estimate; MDS pairs all combined red-cell count with another marker (HGB/RBC $-$0.079). The strongest type~2 diabetes pair, glucose and age ($-$0.066), exceeded the sum of its singletons by 0.017, the largest super-additive excess of any top-ranked pair. Top pairs for the remaining endpoints combined the dominant singletons: age for cardiovascular (all five top pairs), eGFR for fungal and bacterial (four of five each), and INR for metastatic disease (all five, led by creatinine/INR $-$0.041).

Masking reflects both the predictive value of the biology and the model's learned reliance on more frequently measured or correlated markers, and should not be read as intrinsic biomarker importance independent of training.

\subsection{Robustness across subgroups}

On the internal test set, removing sex, age, both, or laboratory imputation each changed the median AUROC by less than 0.005 from the full model (0.740), whereas ensembling contributed more substantially ($-$0.037); on the external MIMIC and MMRF cohorts the same ablations had larger effects and overall discrimination was lower (Figure~\ref{fig:subgroup_ablation}). Subgroup analyses across age and sex are reported in the Supplementary Results.

\begin{figure*}[ht!]
    \centering
    \includegraphics[width=\textwidth]{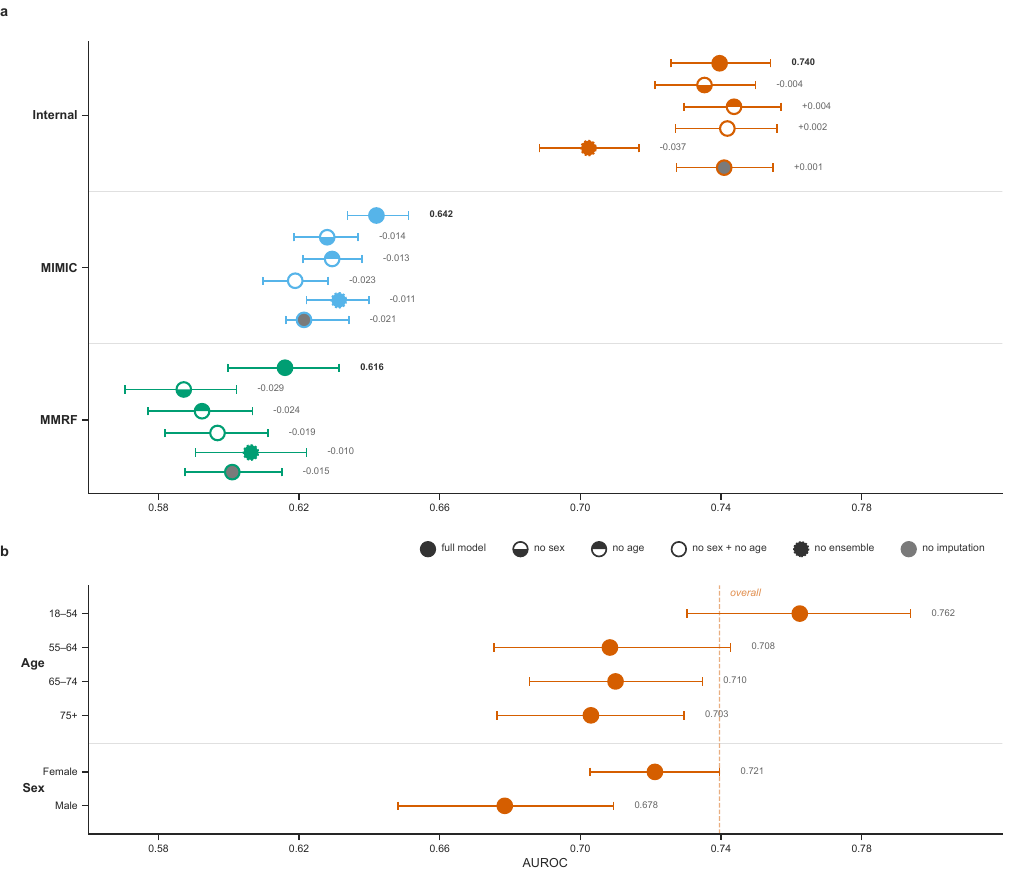}
\caption{\textbf{Subgroup and ablation analyses.}
Mean area under the receiver operating characteristic curve (AUROC) with 95\% bootstrap confidence intervals across subgroups and ablations. For each ICD code, the median over 1000 bootstrap runs was computed; dots show the mean of these per-code medians, restricted to endpoints with $\geq$10 positives per bootstrap. 
\textbf{a}~Performance for the full model and for retraining with sex, age, or both set to zero (value and validity marker). Gray dots indicate retraining without the imputation module, and dotted outlines retraining without model ensembling. Differences are minor on the internal dataset, except for ensembling ($-$0.037), but substantial on both external datasets (MIMIC, MMRF), indicating that demographics and the imputation module improve generalization. \textbf{b}~Performance on the internal dataset stratified by age group and sex. The youngest stratum (18–54) shows the highest AUROC (0.762); the remaining age groups are comparable at a lower level (0.703–0.708). Discrimination is higher for female (0.721) than male (0.678) patients ($\Delta$ = 0.043), yet both fall below the overall AUROC (0.740, dashed line), indicating that the full model's discrimination partly relies on sex differences.}
\label{fig:subgroup_ablation}
\end{figure*}

\subsection{External validation across independent healthcare systems}
External validation was performed on the MIMIC-IV v3.1 database~\cite{PhysioNet-mimiciv-3.1} and the MMRF CoMMpass IA24 cohort~\cite{rustad2019commpass} under out-of-distribution deployment without fine-tuning or cohort-specific recalibration. ICD-10-GM training labels were harmonized to ICD-10-CM and CoMMpass concept-level annotations using a clinically curated crosswalk (Supplementary Methods, Supplementary Table~\ref{tab:label_crosswalk}). Inference was performed across 8{,}150 MIMIC-IV and 7{,}347 CoMMpass sequences.

At the group level (Figure~\ref{fig:subcohort_comparison}b, Table~\ref{tab:performance_comparison_extern_grouped}), transferability was strongest for renal and metabolic endpoints across both cohorts. Kidney disease transferred comparably across MIMIC-IV and CoMMpass (AUROC 0.694 and 0.668, respectively). Individual chronic kidney disease stages, end-stage renal disease and dialysis dependence achieved AUROCs up to 0.85 in MIMIC-IV, while renal and diabetic complication codes reached 0.74--0.76 in CoMMpass. Type~2 diabetes transferred comparably across cohorts (0.650--0.682). MIMIC-IV additionally retained discrimination for metastatic disease (0.667) and myelodysplastic syndrome (0.663), whereas these endpoints were not directly evaluable in CoMMpass because metastatic deposits and MDS are not explicitly represented in its coding structure.

Performance for all-cause mortality, cardiovascular disease, and infectious complications remained moderate across cohorts, while renal and metabolic endpoints showed the strongest transferability. Overall, external validation demonstrated that routine laboratory trajectories capture complication-specific physiological signatures that generalise across independent healthcare systems despite differences in patient populations and coding structure.

\section{Discussion}
Short-term routine laboratory trajectories encode organ-level deterioration that precedes the clinical onset of treatment-associated complications. Predictive signal persisted across extended forecasting horizons for kidney disease, metastatic disease and fungal infections, despite gradual attenuation with increasing temporal distance from diagnosis. The advantage of temporal over cross-sectional modeling was most apparent for endpoints characterized by progressively evolving laboratory abnormalities rather than isolated measurements. The model predicted all eight endpoint groups above prevalence, with discrimination preserved (AUROC 0.71–0.75) for rare endpoints such as MDS and fungal infections where class imbalance typically degrades model performance. The predictions were biologically coherent across two mechanistically distinct cancers, temporally structured beyond what single-timepoint models could capture for most endpoints, and driven by interpretable biomarker signatures aligned with known pathophysiology.

Cardiovascular disease, bacterial infections, and fungal infections generalised across both malignancies with closely comparable discrimination, despite arising through different mechanisms in each cancer. Infection risk in myeloma carries a substantial disease-intrinsic component from humoral immunodeficiency, whereas in ovarian cancer it is predominantly treatment-mediated. The shared predictability reflects not shared biology, but shared laboratory observability. Where discrimination diverged, the gap tracks this same property, not tumour type.

All-cause mortality discriminated better in myeloma than in ovarian cancer (AUROC 0.76 vs 0.63) with the cause-of-death distribution accounting for most of the gap. Myeloma mortality is dominated by infection (immunoparesis~\cite{chahin_clinical_2022}, treatment-related neutropenia~\cite{palumbo_how_2012}), renal failure, and progressive cytopenias~\cite{Mai2018,blimark_risk_2024}, all of which manifest in routine panels. Ovarian cancer mortality is dominated by anatomic complications of peritoneal disease such as bowel obstruction and cachexia~\cite{callaway_mechanisms_2023,ochoa_inoperable_2022,rustin_definitions_2011}, adjudicated by imaging and CA-125 rather than haematology or chemistry. The terminal decline is therefore more laboratory-visible in myeloma than in ovarian cancer.

The metastatic endpoints provided a harder test of the principle as of their different biology. Prevalences of lymph node (C77), respiratory and digestive (C78), and other-site (C79) secondary malignancies differed by an order of magnitude between cohorts, with C78 reaching 27.5\% in ovarian cancer versus 1.2\% in myeloma, reflecting that ovarian cancer spreads by solid-tumour routes~\cite{bayraktar_ovarian_2024} whereas myeloma does not metastasise in the same sense. Group-level discrimination was nevertheless near-identical for C78 (AUROC 0.60 vs 0.61) and C79 (0.60 vs 0.61), indicating that the model captures the laboratory signature of advanced disease burden in both cancers despite the biological difference between them. C77 diverged sharply (myeloma 0.83 vs ovarian 0.60), and the divergence again tracks observability. In myeloma, C77 plausibly captures nodal extramedullary involvement with characteristic cytopenias~\cite{usmani_extramedullary_2012}. In ovarian cancer, C77 captures heterogeneous nodal disease without a distinctive laboratory profile.

Myelodysplastic syndromes diverged sharply in the internal data (OC AUROC 0.93, MM AUROC 0.50), but 21 combined positive cases leave both estimates unstable. MIMIC-IV, with 41 MM cases, returned AUROC 0.66 and indicates a real signal in myeloma that the smaller internal sample missed. The underlying signature, pre-MDS cytopenias, macrocytosis, and elevated red-cell distribution width~\cite{rauw_validation_2011} arising from clonal haematopoiesis and therapy-related marrow injury~\cite{galli_relationship_2021,cai_clonal_2025}, is laboratory-observable by definition. However, if platinum~\cite{Travis1999} and PARP inhibitor~\cite{Tuninetti2024} exposure amplify the signal in ovarian cancer relative to therapy-related marrow injury in myeloma cannot be resolved at this sample size.

The transformer's gain over non-sequential baselines varied from $+$0.11 for myelodysplastic syndromes to zero for metastatic and kidney disease. Because the baselines and the transformer differ only in their capacity to model dependencies across timepoints, the variation tracks how much predictive information each endpoint carries in its trajectory rather than in the static distribution of laboratory values. The endpoints where the transformer gained most are slow-evolving. Myelodysplastic syndromes manifest as progressive cytopenias rather than a single abnormal value. Type~2 diabetes is preceded by glycaemic drift over an extended pre-diagnostic window~\cite{Tabak2012}. For endpoints without a gain, the analysis cannot distinguish a weak temporal signal from one the non-sequential baseline captures equally well.

Together, the baseline and horizon analyses define which complications carry sufficient temporal signal for early risk stratification. For slow-evolving endpoints, the lead time is long enough that preventive intervention has room to act~\cite{lloyd-jones_cardiovascular_2023,Tabak2012}.

Biomarker masking recovered diagnosis-specific signatures consistent with known pathophysiology, despite the model receiving no biological priors. Where a canonical laboratory marker exists, it dominated, namely glucose for type 2 diabetes and the glomerular filtration estimates for kidney disease, whose joint removal produced the largest pairwise effect of any endpoint. Because both estimates derive from creatinine and age, the single-feature drops are lower bounds on the contribution of filtration information. For myelodysplastic syndrome the signal was carried by anaemia (red-cell count and haemoglobin), with red-cell distribution width co-leading the broadly distributed mortality signature in line with its established prognostic association~\cite{Patel2009}. A small set of systemic markers, principally albumin and eGFR, recurred across endpoints, consistent with the model reading shared disease burden alongside endpoint-specific signal, and the strongest analyte pairs exceeded additive expectation, indicating that the model captures trajectory interactions rather than marginal effects. Notably, neutrophil count produced no measurable effect for bacterial infections, which were instead led by markers of chronic systemic deterioration and disease burden (eGFR, albumin, haemoglobin, and red-cell distribution width), consistent with a 730-day horizon targeting cumulative organ dysfunction and immune compromise making the patient more susceptible to infections, rather than acute neutropenic risk~\cite{Raje2022,Nucci2009}.

Age dominated single-feature masking for cardiovascular disease but barely affected ablation. Retraining without sex or age produced negligible change internally ($-$0.004, $+$0.004) and small drops on MIMIC ($-$0.014, $-$0.013), with larger degradation on MMRF ($-$0.029, $-$0.024). MMRF lacks eGFR and eGFR-MDRD, removing the reconstruction path through derived filtration markers. The reliance surfaced by masking thus reflects largely learned redundancy rather than unique information content, recoverable wherever the derived markers are present. Because these drivers are traceable to specific laboratory inputs, model behaviour supports clinical interpretability.

Complication-specific models for nephrotoxicity~\cite{Okawa2022}, febrile neutropenia~\cite{Cho2020}, and cardiotoxicity~\cite{Yagi2024} achieve strong discrimination individually but predict single endpoints from cross-sectional data and cannot be composed into a multi-organ profile. 
General-purpose EHR transformers such as Foresight~\cite{Kraljevic2024}, Delphi-2M~\cite{Shmatko2025}, and CoMET~\cite{CoMET2025} predict thousands of diagnoses but operate on coded events and tokenise laboratory values, requiring richly-coded event streams that most oncology departments do not routinely maintain. 
The closest comparator is SPARC~\cite{Fu2025}, a laboratory-trajectory transformer predicting six endpoints in pan-cancer populations. The present model extends this direction to 162 diagnosis-specific endpoints within defined disease populations. For myelodysplastic syndromes, existing risk models such as CHRS~\cite{Weeks2023} and MN-predict~\cite{Gu2023} achieve strong discrimination but require next-generation sequencing, placing them outside routine clinical monitoring. Both nonetheless found routine CBC parameters carried independent predictive value~\cite{Weeks2023,Gu2023}, consistent with the masking analysis reported here. 

Predictive enrichment for myelodysplastic syndromes could identify high-risk patients for intensified haematological surveillance or emerging prevention strategies, such as CDK4/6 inhibition to limit chemotherapy-induced expansion of TP53-mutant clonal haematopoiesis~\cite{Chan2026}, in a setting where monitoring currently relies on periodic blood counts without systematic risk stratification~\cite{Tuninetti2024}. This requires no testing beyond standard oncological care. However, enrichment over prevalence demonstrates that predictive signal exists but does not establish the discrimination needed for individual management; translating it into actionable risk scores will require prospective calibration studies defining decision thresholds for specific interventions.

Cross-system external validation tested transfer to two structurally distinct cohorts without fine-tuning or recalibration. Type~2 diabetes transferred most stably, matching internal discrimination on MIMIC-IV (AUROC 0.65) and exceeding it on MMRF CoMMpass (0.68), while all-cause mortality and cardiovascular disease were largely retained on MMRF despite attenuation on MIMIC-IV. On MIMIC-IV, myelodysplastic syndrome and metastatic disease were best preserved, with MDS retaining 2.58-fold enrichment over prevalence. Infection endpoints degraded most on MIMIC-IV (bacterial 0.56, fungal 0.56) but recovered partially on MMRF (0.60, 0.62). Across both cohorts, all evaluable endpoints maintained enrichment at or above prevalence, indicating transferable laboratory signal even where discrimination was attenuated. 
The variable degradation reflected at least three structural coding differences between ICD-10-GM and the target systems, including obligatory versus optional secondary diagnosis coding for pathogens, non-equivalent cardiovascular and renal disease subcategorisation, and divergent reimbursement incentives that drive differential capture of comorbidities across DRG systems~\cite{jette2010,busse2011}. Disentangling these effects from genuine cross-population shift is not possible with cross-sectional ICD coding alone, and the residual question of how much each contributes is a structural problem for every ICD-coded prediction model deployed across national systems.

This study has limitations. No prospective validation has been performed. Although the model was developed at a single center, the identified laboratory signatures transferred to two independent external cohorts (MIMIC-IV and MMRF CoMMpass), supporting generalizability beyond the development population. The model receives no treatment information, and the observed associations may therefore reflect disease progression, comorbidity burden, or treatment effect. Disentangling these contributions will require models that integrate treatment exposure data. The steeper decline in discrimination for bacterial infections across forecasting horizons suggests that their predictive signal is concentrated closer to clinical onset. Accordingly, the 730-day prediction horizon is appropriate for gradually evolving endpoints but may limit clinical actionability for acute infectious complications, where shorter horizons and more frequent model updates would constitute a distinct modeling task. The reliance on ICD-10-coded diagnoses as outcome labels introduces potential misclassifications, as demonstrated by the structural coding differences identified in the cross-system validation, and coding accuracy, completeness, and temporal assignment of diagnosis dates vary across institutions and over time. The 25-year observation window spans multiple changes in laboratory assays, treatment standards, and coding practices that may introduce confounders not captured by the current modeling framework. PARP inhibitors entered routine first-line use in the final five to seven years of the observation period. Given the latency between exposure and therapy-related myelodysplasia, the MDS signal in the ovarian subcohort more plausibly reflects platinum toxicity. Characterising PARP-associated myelodysplasia~\cite{Morice2021} would require cohorts with longer, better-characterised exposure.


In conclusion, routine laboratory trajectories collected during standard cancer treatment encode complication-specific organ deterioration weeks to months before clinical onset. The signal sits inside data that every oncology unit already produces and every hospital information system already stores, making laboratory trajectory modeling the cheapest dense longitudinal monitoring channel available in oncology, with no new assay, sequencing run, or imaging study required to deploy it. The same trajectory is the natural substrate for patient-level digital twins of cancer, where a continuously updating organ-state representation is required and no competing modality delivers comparable temporal density at comparable cost. The 162-endpoint resolution shown here demonstrates that this substrate carries complication-specific rather than purely prognostic information, with interpretable biomarker signatures that keep the predictions auditable rather than opaque.

\section*{Methods}

\subsection{Study cohort and clinical data}
A cohort of 2,328 patients with multiple myeloma treated at a tertiary care centre (Klinikum rechts der Isar, Technical University of Munich) between January 1st 2000 and January 1st 2026 was combined with 2,141 patients with ovarian cancer from the same institution. For both subcohorts, laboratory measurements and ICD-10-coded diagnoses used as input to the diagnosis prediction model were obtained from the hospital information system. Of these 4,469 patients, 4,236 had valid laboratory measurements recorded. After restricting analytes to 39 routinely tested parameters, 4,184 patients were retained; subsequent removal of patients with more than 75\% missing values across retained analytes yielded the final cohort of 3,905 patients used for diagnosis prediction (Figure~\ref{fig:health_forecaster}b). Independently, for the myeloma subcohort, a five-stage LLM-based information extraction pipeline using locally deployed Llama 3.3-70B~\cite{Grattafiori2024} was applied to 90,036 unstructured German-language clinical documents, including 31,790 internal discharge summaries and specialised reports from radiology, pathology, and nuclear medicine departments. This pipeline produced a structured myeloma dataset encompassing treatment lines, cytogenetic assessments, bone marrow infiltration, refractory status, and clinical endpoints (Supplementary Methods, Extended Data Fig.~\ref{fig:structuring_overview}, Supplementary Tables~\ref{tab:entity_accuracy_grouped_N},~\ref{tab:entity_external_accuracy_doc_level}). This dataset contextualises the cohort but was not used as model input. For the ovarian cancer subcohort, equivalent clinical metadata were available directly from the hospital information system.

\subsection{Diagnosis prediction from laboratory trajectories}

\paragraph{Dataset construction.}

The initial dataset comprised 4,094,479 laboratory measurements across 4,236 patients and 272 analytes. Input features were restricted to 39 analytes routinely obtained during standard clinical assessment, spanning haematology, metabolic, hepatic, coagulation, inflammatory, and endocrine panels (Extended Data Table~\ref{tab:lab_inputs}), rather than tests ordered in response to specific clinical suspicion. This restriction reduces indication bias and limits predictions to measurements available in routine clinical care regardless of institutional infrastructure. After analyte filtering and removal of patients with more than 75\% missing values, 2,777,595 measurements across 3,905 patients remained (Figure~\ref{fig:health_forecaster}b).

Patient laboratory trajectories were segmented into non-overlapping temporal intervals with durations sampled uniformly between 10 and 30 days. Laboratory values obtained on the same calendar day were averaged per analyte to mitigate indication bias, as same-day repeat measurements typically reflect acute clinical concern rather than the gradual physiological trends targeted by the 730-day prediction horizon. Each day containing at least one valid measurement was represented as a single input token encoding the 39 analyte values together with the patient's age in days. A binary sex token was appended once per sequence. Intervals without measurements were discarded. The number of sequences per patient was capped at ten to limit overrepresentation of heavily monitored individuals, with the cap increased to 20 for the myelodysplastic syndrome endpoint to compensate for its low prevalence (Supplementary Methods). To prevent label leakage, sequences were excluded on a per-diagnosis basis if the target diagnosis had been documented prior to the first day following the end of the laboratory observation window.

Clinical outcomes were defined using 162 ICD-10 codes curated by a board-certified oncologist from approximately 3,000 distinct codes documented across the cohort. Each diagnosis was required to occur in at least 50 patients across the full dataset. Diagnosis prediction was formulated as a multi-label classification task estimating the probability of each diagnosis within 730 days following the end of the laboratory observation window. This horizon captures complications that manifest across one to two lines of therapy in both diseases~\cite{Rajkumar2022,Tuninetti2024} (see Supplementary Methods for detailed justification). The final cohort was split at the patient level into training (3,125 patients; 12,145 sequences), validation (390 patients; 1,453 sequences), and test (390 patients; 1,579 sequences) sets.

\paragraph{Laboratory value imputation.}

Routine laboratory panels rarely measure all 39 analytes simultaneously, and the pattern of missingness may itself encode clinical context. A dedicated transformer-based imputation model was therefore trained prior to diagnosis prediction. The imputation model produced dense timestep representations, preventing the prediction model from exploiting differential test-ordering patterns rather than physiological values. The imputation model followed a masked prediction objective in which individual analytes were randomly masked with probability 0.2 and entire timesteps with probability 0.05. The model was trained to reconstruct masked values using mean squared error loss (architectural details and hyperparameters in Supplementary Table~\ref{tab:imputation_architecture}).

\paragraph{Prediction model.}

Temporal dependencies across laboratory timesteps were modelled using a bidirectional transformer encoder adopting the architectural components of Qwen-3~\cite{yang2025}, including rotary positional embeddings, SwiGLU activations, RMSNorm, and grouped-query attention. These design choices reflect current best practices for capturing long-range dependencies in sequential data and have been validated extensively at scale in the language domain. The same inductive biases, particularly the capacity to model complex interactions across ordered sequences, were hypothesized to transfer to longitudinal laboratory trajectories. All weights were trained from scratch on structured laboratory data without language pretraining. Bidirectional attention was chosen because the prediction task operates on completed observation windows rather than generating sequential, token-wise predictions. Each laboratory interval is fully observed before diagnosis probabilities are computed, and bidirectional context allows the model to interpret early measurements in light of later values within the same interval. For each interval, normalised and imputed laboratory values were projected into a continuous embedding space via a linear input layer. A linear prediction head mapped the attention-pooled final-layer representations of the transformer to multi-label diagnosis probabilities. Models were trained using weighted focal loss with class weights derived from training set diagnosis frequencies, and optimised with Adam using cosine annealing (architectural details and hyperparameters in Supplementary Table~\ref{tab:prediction_architecture}). The pipeline is illustrated in Figure~\ref{fig:health_forecaster}a.

\paragraph{Evaluation.}

The combined training and validation sets were partitioned into five patient-level folds. One model was trained per fold, with four folds used for training and the held-out fold for validation. Hyperparameters were selected based on mean area under the receiver operator curve (AUROC) across validation folds. For final evaluation on the held-out test set, predictions were obtained by averaging raw logits across all five fold-specific models before applying the sigmoid function.

AUROC was chosen as the primary discrimination metric because it is threshold-independent, comparable across endpoints with differing prevalences, and widely reported in the literature, facilitating comparison with existing models. To contextualise discrimination relative to baseline event rates, predictive enrichment was quantified as the ratio of average precision to endpoint prevalence (enrichment ratio). Average precision is additionally reported for all endpoints in Supplementary Tables~\ref{tab:performance_comparison} and~\ref{tab:performance_comparison_grouped}, as it directly reflects the clinical use case of identifying affected patients under class imbalance~\cite{saito2015}.

Group-level predictions were constructed by taking the maximum predicted probability across constituent diagnoses. Confidence intervals were obtained from 1000 nonparametric bootstrap resamples of the test set (Supplementary Methods).

\paragraph{Baselines.}

To isolate the effect of temporal modeling, the transformer was compared against two non-sequential baselines, logistic regression (linear) and XGBoost (non-linear), following the baseline methodology described in the CoMET benchmark~\cite{CoMET2025}. Each laboratory sequence was converted into a bag-of-words count vector in which each feature is the number of measurements falling in a given (analyte, value-bin) cell, with values binned into deciles per analyte. Logistic regression features were scaled to unit maximum absolute value on the training set; XGBoost, being invariant to monotonic per-feature rescaling, consumed raw counts. Both baselines received the same demographic features (age and sex) and training labels as the transformer. XGBoost was included as the strongest supervised baseline, as it matched or exceeded CoMET-L on incident disease prediction tasks with low endpoint prevalence in that benchmark. This representation preserves each analyte's binned value distribution over the window but discards the order and timing of measurements, so the transformer's improvement over these baselines quantifies the predictive information carried by the temporal trajectory beyond an order-free summary of the same values.

\paragraph{External validation.}

Cross-system transferability was assessed on two structurally distinct external cohorts. First, MIMIC-IV (version 3.1), a publicly available critical care database from Beth Israel Deaconess Medical Center, Boston, USA. Patients were selected by ICD-10-CM diagnosis codes corresponding to multiple myeloma (C90.0x) or ovarian cancer (C56.x), yielding 1,626 patients (930 multiple myeloma, 697 ovarian cancer; one patient carried both diagnoses) comprising 8,150 sequences. Laboratory values for the same 39 analytes used in the primary analysis were extracted and mapped to their MIMIC-IV equivalents using manually curated crosswalks (Supplementary Table~\ref{tab:lab_inputs}). Two of the 39 analytes (plateletcrit and platelet distribution width) were unavailable in MIMIC-IV and were treated as missing inputs throughout that cohort. Diagnostic endpoints were mapped from the 162 ICD-10-GM targets to ICD-10-CM codes through manually curated crosswalks verified against the codes present in MIMIC-IV (Supplementary Table~\ref{tab:label_crosswalk}). No official crosswalk exists between ICD-10-GM and ICD-10-CM, as both are independent national modifications of WHO ICD-10~\cite{jette2010}; mappings were therefore curated by clinical review. Second, the MMRF CoMMpass IA24 release, a prospective longitudinal cohort of newly diagnosed multiple myeloma patients with serial laboratory monitoring, comprising 1,143 multiple myeloma patients yielding in 7,347 sequences. Laboratory values for the same 39 analytes were extracted and mapped to MMRF's native laboratory concepts using manually curated crosswalks (Supplementary Table~\ref{tab:lab_inputs}); unavailable analytes were treated as missing inputs throughout. Diagnostic endpoints were harmonised from the 162 ICD-10-GM targets to MMRF's concept-level coding through manually curated mappings (Supplementary Table~\ref{tab:label_crosswalk}); metastatic codes and myelodysplastic syndrome had no equivalent concept, and were excluded from evaluation. For both cohorts, evaluation was performed on the entire dataset, and group-level and per-endpoint AUROC were computed using the same procedures as for the internal test set.

\paragraph{Biomarker importance analysis.}
 
Diagnosis-specific laboratory signatures were quantified through a 
leave-one-out masking approach. For each of the 39 analytes and age, 
and for each diagnostic group, the feature value and its 
corresponding validity indicator were set to zero across all 
timesteps in the test set, distinguishing masked entries from true 
zero measurements. The change in AUROC relative to the unmasked 
baseline was recorded. Pairwise masking was performed analogously 
for all $\binom{40}{2} = 780$ feature pairs to identify synergistic 
biomarker combinations.

\paragraph{Software and hardware.}

All models were implemented in Python using PyTorch 2.8.0 and Hugging Face Transformers 4.57.0. Evaluation metrics were computed using scikit-learn 1.7.2. Training was performed on a single Nvidia H100 GPU.

\section*{Data availability}
The clinical dataset underlying this study contains protected health information and cannot be shared publicly under applicable data protection regulations (EU General Data Protection Regulation).

\section*{Code availability}
Source code for the structuring, imputation and prediction model 
training and evaluation, together with trained model weights, will 
be made available at \url{https://jalu1870.github.io/lab-forecast/} 
upon publication.

\section*{Acknowledgements}
This work is supported by the Bavarian State Ministry of Health, Care and Prevention through the GO-TWIN project (grant number DGP-2024-01).

\section*{Author contributions}
J.L. conceived the study, planned and conducted all experiments for clinical data structuring and diagnosis prediction, constructed the information extraction pipeline, designed the model evaluation framework and statistical analyses, performed input preprocessing, and wrote the manuscript. K.Bra. reviewed the manuscript, provided clinical expertise on multiple myeloma, contributed to endpoint definition for clinical data structuring, and selected diagnoses for prediction. J.La. reviewed the manuscript and provided clinical expertise on ovarian cancer. C.Wi. provided access to laboratory data. F.G. performed data annotations and advised on the design of the structuring study. T.L. and C.Z. performed data annotations. M.G. contributed expertise on multiple myeloma endpoint definition. F.P., A.Z., H.H., A.N., F.D., Z.B.C., J.M., S.H.K., S.Z., 
F.B., M.H., J.K., and M.M. critically reviewed the manuscript and 
provided feedback. C.Wa. supervised the project and recommended the XGBoost comparison analysis. L.A. and K.Bre. jointly supervised this work. They provided computational resources, advised on experimental design, figure design, and template construction, and contributed to manuscript editing and revision.

\section*{Competing interests}
FB received honoraria from Amgen, Johnson \& Johnson, Bristol Myers Squibb (BMS), AbbVie, and GSK, travel support from Amgen, Johnson \& Johnson, and BMS, and served on advisory boards for Amgen, Johnson \& Johnson, BMS, AbbVie, and GSK. MH received honoraria from Johnson \& Johnson, Sanofi, GSK, Oncopeptides, and Pfizer, payment for expert testimony from Johnson \& Johnson and Oncopeptides, travel support from Johnson \& Johnson, Oncopeptides, Pfizer, and Amgen, and served on advisory boards for Johnson \& Johnson, Sanofi, Oncopeptides, and Pfizer. All other authors declare no competing interests.

\section*{Additional information}
Correspondence and requests for materials should be addressed to
Jannik L\"ubberstedt (\url{jannik.luebberstedt@tum.de}).

\bibliographystyle{unsrt}
\bibliography{sample}

\clearpage
\onecolumn

\section{Extended Data}

\begin{table}[ht!]
\small
\centering
\caption{Comparison of multiple myeloma datasets by available clinical content and data granularity.}
\label{tab:mm_datasets_comparison}
\renewcommand{\arraystretch}{1.2}
\resizebox{\textwidth}{!}{%
\begin{tabular}{@{} l l l l l r @{}}
\toprule
\textbf{Dataset} &
\textbf{Comorbidities} &
\textbf{Treatment} &
\textbf{Medications} &
\textbf{Labs / BM / Imaging / Pathology} &
\textbf{Patients} \\
\midrule
MMRF CoMMpass~\cite{rustad2019commpass} &
Limited &
Detailed &
MM-specific only &
Labs, BM, cytogenetics &
1,143 \\
Connect MM Registry~\cite{stephens2019connectmm} &
Moderate &
High-level &
MM-specific only &
Baseline labs &
3,011 \\
SEER~\cite{seer_overview} &
None &
Initial only, coded &
None &
None &
population registry \\
This study &
Detailed, narrative &
Longitudinal, narrative &
Complete &
Comprehensive, narrative &
2,328 \\
\bottomrule
\end{tabular}%
}
\end{table}

\begingroup
\setlength{\LTleft}{\fill}
\setlength{\LTright}{\fill}
\begin{longtable}{@{} p{4cm} p{5cm} r r @{}}
\caption{\textbf{Laboratory parameters included in the analysis and their MIMIC-IV and CoMMpass MMRF equivalents.}
Each of the 39 routinely tested analytes used as model inputs was matched to its corresponding entry in the MIMIC-IV \texttt{d\_labitems} table by clinical equivalence. Two analytes (plateletcrit and platelet distribution width) have no corresponding item in MIMIC-IV and were treated as missing inputs to the imputation model throughout the external cohort. The CoMMpass column lists the corresponding analyte code in the MMRF CoMMpass dataset where available.}
\label{tab:lab_inputs} \\
\toprule
\textbf{Category} & \textbf{Laboratory Parameter} & \textbf{MIMIC itemid} & \textbf{CoMMpass} \\
\midrule
\endfirsthead
\toprule
\textbf{Category} & \textbf{Laboratory Parameter} & \textbf{MIMIC itemid} & \textbf{CoMMpass} \\
\midrule
\endhead
\midrule
\multicolumn{4}{r@{}}{\emph{Continued on next page}} \\
\endfoot
\bottomrule
\endlastfoot
\multirow{17}{*}{\textbf{Complete Blood Count (CBC)}    }
 & Basophils & 51146 & --- \\*
 & Eosinophils & 51200 & --- \\*
 & Hematocrit (Hct) & 51221 & --- \\*
 & Hemoglobin (Hb) & 51222 & HGB \\*
 & Lymphocytes & 51244 & --- \\*
 & Mean Corpuscular Hemoglobin & 51248 & --- \\*
 & Mean Corpuscular Hem.\ Conc. & 51249 & --- \\*
 & Mean Corpuscular Volume & 51250 & --- \\*
 & Monocytes & 51254 & --- \\*
 & Mean Platelet Volume & 52142 & --- \\*
 & Neutrophils & 51256 & NEUT \\*
 & Plateletcrit & --- & --- \\*
 & Platelet Distribution Width & --- & --- \\*
 & Platelet Count & 51265 & PLT \\*
 & Red Blood Cell Count (RBC) & 51279 & --- \\*
 & Red Cell Distribution Width & 51277 & --- \\*
 & White Blood Cell Count (WBC) & 51301 & WBC \\*
\midrule
\multirow{8}{*}{\textbf{Electrolytes \& Chemistry}}
 & Blood Urea Nitrogen & 51006 & BUN \\
 & Creatinine & 50912 & CREAT \\
 & Calcium (Ca$^{2+}$) & 50893 & CA \\
 & Blood Glucose & 50931 & GLU \\
 & Potassium (K$^+$) & 50971 & --- \\
 & Sodium (Na$^+$) & 50983 & --- \\
 & Anion Gap & 50868 & --- \\
 & Uric Acid & 51007 & --- \\
\midrule
\multirow{2}{*}{\textbf{Kidney Function}}
 & Estim.\ GFR (CKD-EPI) & 53161 & --- \\
 & Estim.\ GFR (MDRD Formula) & 50920 & --- \\
\midrule
\multirow{7}{*}{\textbf{Liver Function}}
 & Serum Albumin & 50862 & ALB \\
 & Alkaline Phosphatase (ALP) & 50863 & --- \\
 & Alanine Aminotransferase (ALT) & 50861 & --- \\
 & Aspartate Aminotransferase (AST) & 50878 & --- \\
 & Gamma-Glutamyl Transferase & 50927 & --- \\
 & Total Bilirubin & 50885 & --- \\
 & Total Protein & 50976 & TP \\
\midrule
\multirow{1}{*}{\textbf{Inflammation}}
 & C-Reactive Protein & 50889 & CRP \\
\midrule
\multirow{3}{*}{\textbf{Coagulation}}
 & International Normalized Ratio & 51237 & --- \\
 & Prothrombin Time (PT) & 51274 & --- \\
 & Activ.\ Partial Throm.\ Time & 51275 & --- \\
\midrule
\multirow{1}{*}{\textbf{Thyroid Function}}
 & Thyroid-Stimulating Hormone & 50993 & --- \\
\end{longtable}
\endgroup

\rowcolors{2}{white}{gray!8}
\setlength{\LTleft}{\fill}
\setlength{\LTright}{\fill}
\begin{longtable}{@{} p{1.7cm} p{4.6cm} p{4.4cm} p{4.4cm} @{}}
\caption{\textbf{Endpoint label crosswalk from ICD-10-GM training labels to external-cohort definitions.}
For each ICD-10-GM endpoint used in training, the table lists the ICD-10-CM target codes used for MIMIC-IV validation and the corresponding MMRF CoMMpass concept used for CoMMpass validation. GM-specific axes not preserved in ICD-10-CM (NYHA class, coronary vessel count, decompensation status, KDIGO stage) were dropped during mapping and instead a closest match of available codes was applied to enable appropriate group-wise comparisons. }
\label{tab:label_crosswalk} \\
\hiderowcolors
\toprule
\textbf{ICD-10-GM} & \textbf{GM description} & \textbf{ICD-10-CM (MIMIC-IV)} & \textbf{MMRF CoMMpass concept} \\
\midrule
\endfirsthead
\hiderowcolors
\toprule
\textbf{ICD-10-GM} & \textbf{GM description} & \textbf{ICD-10-CM (MIMIC-IV)} & \textbf{MMRF CoMMpass concept} \\
\midrule
\endhead
\midrule
\multicolumn{4}{r@{}}{\emph{Continued on next page}} \\
\endfoot
\bottomrule
\endlastfoot
\showrowcolors

A04.7 & C. difficile enterocolitis & A04.7, A04.71, A04.72 & C. difficile infection \\
A04.70 & C. difficile enterocolitis & A04.7, A04.71, A04.72 & C. difficile infection \\
A09 & Gastroenteritis, infectious & A09 & Infectious gastroenteritis \\
A09.0 & Gastroenteritis, infectious & A09 & Infectious gastroenteritis \\
A09.9 & Gastroenteritis, unspecified & A09 & Infectious gastroenteritis \\
A41.1 & Sepsis, Staphylococcus & A41.1 & Sepsis \\
A41.9 & Sepsis, unspecified & A41.9 & Sepsis \\
A41.51 & Sepsis, E. coli & A41.51 & Sepsis \\
A41.58 & Sepsis, other Gram-negative & A41.59 & Sepsis \\
B00.9 & Herpesviral infection, unspecified & B00.9 & Herpes simplex \\
B02.2 & Zoster w/ nervous-system involvement & B02.2, B02.21, B02.29 & Herpes zoster \\
B02.9 & Herpes zoster, uncomplicated & B02.9 & Herpes zoster \\
B18.1 & Chronic viral hepatitis B & B18.1 & Chronic hepatitis B/C \\
B18.2 & Chronic viral hepatitis C & B18.2 & Chronic hepatitis B/C \\
B24 & HIV disease, unspecified & B20, B24 & [unmappable] \\
B25.88 & Other cytomegaloviral disease & B25.8 & [unmappable] \\
B27.0 & Infectious mononucleosis, EBV & B27.0, B27.00, B27.01, B27.02, B27.09 & EBV mononucleosis \\
B35.1 & Tinea unguium & B35.1 & Onychomycosis / Fungal infection \\
B37.0 & Candidal stomatitis & B37.0 & Candidiasis / Fungal infection \\
B37.88 & Candidiasis, other sites & B37.8, B37.81, B37.82, B37.83, B37.84, B37.89 & Candidiasis / Fungal infection \\
B44.1 & Pulmonary aspergillosis & B44.1 & Aspergillosis / Fungal infection \\
B47.9 & Mycetoma, unspecified & B47.9 & Fungal infection \\
B95.2 & Streptococcus grp D / Enterococcus & B95.2 & Other bacterial infection \\
B95.6 & Staphylococcus aureus & B95.6, B95.61, B95.62 & Strep./staph. bacteraemia \\
B95.7 & Other staphylococci & B95.7 & Strep./staph. bacteraemia \\
B95.8 & Unspecified staphylococci & B95.8 & Strep./staph. bacteraemia \\
B95.48 & Other specified streptococci & B95.4 & Strep./staph. bacteraemia \\
B95.90 & Other gram-positive aerobes & B96.8, B96.89 & Strep./staph. bacteraemia \\
B96.2 & E. coli / Enterobacterales & B96.2, B96.20, B96.21, B96.22, B96.23, B96.29 & Other bacterial infection \\
B96.5 & Pseudomonas & B96.5 & Other bacterial infection \\
B96.6 & Bacteroides fragilis & B96.6 & Other bacterial infection \\
B96.8 & Other specified bacterial agents & B96.8, B96.81, B96.82, B96.83, B96.89 & Other bacterial infection \\
B99 & Other infectious disease, unspec. & B99, B99.8, B99.9 & Unspecified infection \\
C73 & Malig. neoplasm, thyroid gland & C73 & Thyroid cancer (SPM if post-MM) \\
C77.0 & Sec. malig., head/neck LN & C77.0 & [construct shift] \\
C77.1 & Sec. malig., thoracic LN & C77.1 & [construct shift] \\
C77.2 & Sec. malig., intra-abdominal LN & C77.2 & [construct shift] \\
C77.3 & Sec. malig., axillary/UE LN & C77.3 & [construct shift] \\
C77.4 & Sec. malig., inguinal/LE LN & C77.4 & [construct shift] \\
C77.5 & Sec. malig., intrapelvic LN & C77.5 & [construct shift] \\
C77.8 & Sec. malig., LN multiple regions & C77.8 & [construct shift] \\
C77.9 & Sec. malig., LN unspecified & C77.9 & [construct shift] \\
C78.0 & Sec. malig., lung & C78.0, C78.00, C78.01, C78.02 & [construct shift] \\
C78.1 & Sec. malig. neoplasm, mediastinum & C78.1 & [construct shift] \\
C78.2 & Sec. malig., pleura & C78.2 & [construct shift] \\
C78.4 & Sec. malig., small intestine & C78.4 & [construct shift] \\
C78.5 & Sec. malig., large intestine & C78.5 & [construct shift] \\
C78.6 & Sec. malig., peritoneum & C78.6 & [construct shift] \\
C78.7 & Sec. malig., liver & C78.7 & [construct shift] \\
C78.8 & Sec. malig., other digestive & C78.8, C78.80, C78.89 & [construct shift] \\
C79.2 & Sec. malig., skin & C79.2 & [construct shift] \\
C79.3 & Sec. malig., brain/meninges & C79.3, C79.31, C79.32 & [construct shift] \\
C79.4 & Sec. malig., nervous system & C79.4, C79.40, C79.49 & [construct shift] \\
C79.5 & Sec. malig., bone/marrow & C79.5, C79.51, C79.52 & [construct shift] \\
C79.6 & Sec. malig., ovary & C79.6, C79.60, C79.61, C79.62, C79.63 & [construct shift] \\
C79.7 & Sec. malig., adrenal gland & C79.7, C79.70, C79.71, C79.72 & [construct shift] \\
C79.9 & Sec. malig., unspecified site & C79.9 & [construct shift] \\
C79.82 & Sec. malig., genital organs & C79.82 & [construct shift] \\
C79.88 & Sec. malig., other sites & C79.89 & [construct shift] \\
D46.2 & Refractory anaemia w/ excess blasts & D46.2, D46.20, D46.21, D46.22 & [unmappable] \\
D46.7 & Other myelodysplastic syndromes & D46.Z & [unmappable] \\
D46.9 & Myelodysplastic syndrome, unspec. & D46.9 & [unmappable] \\
E03.8 & Other hypothyroidism & E03.8 & Hypothyroidism \\
E03.9 & Hypothyroidism, unspecified & E03.9 & Hypothyroidism \\
E04.2 & Nontoxic multinodular goitre & E04.2 & Goitre/thyroid nodule \\
E04.9 & Nontoxic goitre, unspecified & E04.9 & Goitre/thyroid nodule \\
E05.0 & Hyperthyroidism, Graves' & E05.00, E05.01 & Hyperthyroidism \\
E05.9 & Hyperthyroidism, unspecified & E05.90, E05.91 & Hyperthyroidism \\
E06.3 & Autoimmune thyroiditis (Hashimoto) & E06.3 & Thyroid disorder \\
E11.20 & T2DM with kidney complications & E11.21, E11.22, E11.29 & Diabetes / Chronic kidney disease \\
E11.90 & T2DM w/o complications & E11.9 & Diabetes \\
E11.91 & T2DM w/o complications (decompensated) & E11.9 & Diabetes \\
E13.90 & Other specified diabetes mellitus & E13.9 & Diabetes \\
F32.0 & Mild depressive episode & F32.0 & Depression \\
F32.1 & Moderate depressive episode & F32.1 & Depression \\
F32.2 & Severe depression w/o psychosis & F32.2 & Depression \\
F32.8 & Other depressive episodes & F32.89 & Depression \\
F32.9 & Depressive episode, unspecified & F32.9, F32.A & Depression \\
F33.1 & Recurrent depression, moderate & F33.1 & Depression \\
F33.2 & Recurrent depression, severe & F33.2 & Depression \\
I10 & Essential hypertension (parent) & I10 & Hypertension \\
I10.00 & Essential hypertension, benign & I10 & Hypertension \\
I10.01 & Malignant essential hypertension & I10 & Hypertension \\
I10.90 & Essential hypertension, no crisis & I10 & Hypertension \\
I10.91 & Essential hypertension, w/ crisis & I10 & Hypertension \\
I11.00 & Hypertensive heart disease, w/ HF & I11.0 & Hypertension / Heart failure \\
I11.90 & Hypertensive heart disease, no HF & I11.9 & Hypertension \\
I12.00 & Hypertensive renal disease, w/ failure & I12.0, I12.9 & Hypertension / Chronic kidney disease \\
I15.10 & Renovascular hypertension & I15.0 & Hypertension \\
I20.0 & Unstable angina & I20.0 & Angina \\
I20.8 & Other angina pectoris & I20.8 & Angina \\
I21.4 & Acute subendocardial MI (NSTEMI) & I21.4 & Myocardial infarction \\
I25.9 & Chronic ischaemic HD, unspecified & I25.9 & Coronary artery disease \\
I25.11 & Atherosclerotic HD, 1-vessel & I25.10, I25.11x & Coronary artery disease / Angina \\
I25.12 & Atherosclerotic HD, 2-vessel & I25.10, I25.11x & Coronary artery disease \\
I25.13 & Atherosclerotic HD, 3-vessel & I25.10, I25.11x & Coronary artery disease \\
I25.19 & Atherosclerotic HD, unspecified & I25.10, I25.11x & Coronary artery disease \\
I25.22 & Old myocardial infarction & I25.2 & Coronary artery disease \\
I42.0 & Dilated cardiomyopathy & I42.0 & Cardiomyopathy \\
I48.0 & Atrial fibrillation, paroxysmal & I48.0 & Atrial fibrillation/flutter \\
I48.1 & Atrial fibrillation, persistent & I48.1, I48.11, I48.19 & Atrial fibrillation/flutter \\
I48.2 & Atrial fibrillation, chronic & I48.2, I48.20, I48.21, I48.91 & Atrial fibrillation/flutter \\
I48.9 & Atrial fibrillation, unspecified & I48.9, I48.91, I48.92 & Atrial fibrillation/flutter \\
I48.10 & Persistent AF (non-standard) & I48.1, I48.19 & Atrial fibrillation/flutter \\
I48.11 & Persistent AF, longstanding (non-std) & I48.1, I48.11 & Atrial fibrillation/flutter \\
I48.19 & Persistent AF, other (non-std) & I48.1, I48.19 & Atrial fibrillation/flutter \\
I50.01 & Secondary right heart failure & I50.81, I50.9 & Heart failure \\
I50.9 & Heart failure, unspecified & I50.9 & Heart failure \\
I50.12 & Left heart failure, NYHA II & I50.1, I50.2x, I50.3x, I50.4x & Heart failure \\
I50.13 & Left heart failure, NYHA III & I50.1, I50.2x, I50.3x, I50.4x & Heart failure \\
I50.14 & Left heart failure, NYHA IV & I50.1, I50.2x, I50.3x, I50.4x & Heart failure \\
I80.1 & Phlebitis/thrombosis, femoral vein & I80.1, I82 & Deep/superficial vein thrombosis \\
I80.2 & Phlebitis, deep LE veins & I80.2, I82 & Deep/superficial vein thrombosis \\
I80.3 & Phlebitis, lower extremity (incl. DVT NOS) & I80.3, I82 & Deep/superficial vein thrombosis \\
I80.8 & Phlebitis, other localisation & I80.8, I80.89 & Deep/superficial vein thrombosis \\
I80.28 & Phlebitis, other deep LE veins & I80.29, I82 & Deep/superficial vein thrombosis \\
I80.80 & Phlebitis, superficial UE veins & I80.8, I80.89 & Deep/superficial vein thrombosis \\
I80.81 & Phlebitis, deep UE veins & I80.8, I80.89, I82.A1 & Deep/superficial vein thrombosis \\
I80.88 & Phlebitis, other localisation & I80.8, I80.89, I80.9, I82.90, I82.91 & Deep/superficial vein thrombosis \\
I82.9 & Embolism/thrombosis, unspec. veins & I82.90, I82.91 & Embolism/thrombosis \\
J96.0 & Acute respiratory failure & J96.0, J96.00, J96.01, J96.02 & Respiratory failure \\
J96.00 & Acute respiratory failure, type I & J96.01 & Respiratory failure \\
J96.9 & Respiratory failure, unspecified & J96.9, J96.90, J96.91, J96.92 & Respiratory failure \\
J96.09 & Acute respiratory failure, unspec. & J96.00 & Respiratory failure \\
K92.2 & GI haemorrhage, unspecified & K92.2 & GI haemorrhage \\
M80.88 & Osteoporosis w/ pathologic fracture & M80.88 & Pathologic/compression fracture / Osteoporosis/osteopenia \\
M81.89 & Other osteoporosis w/o fracture & M81.8 & Osteoporosis/osteopenia \\
M81.99 & Osteoporosis, unspecified & M81.9 & Osteoporosis/osteopenia \\
M89.50 & Osteolysis, multi-site / unspec. & M89.50, M89.59 & Lytic bone lesion \\
M89.55 & Osteolysis, femur & M89.55, M89.551, M89.552, M89.559 & Lytic bone lesion \\
M89.58 & Osteolysis, other site & M89.58 & Lytic bone lesion \\
M89.59 & Osteolysis, unspecified site & M89.50, M89.59 & Lytic bone lesion \\
N17.8 & Other acute kidney failure & N17.8 & Acute kidney injury \\
N17.9 & AKI, unspecified & N17.9 & Acute kidney injury \\
N17.91 & AKI, stage 1 & N17.9 & Acute kidney injury \\
N17.92 & AKI, KDIGO stage 2 & N17.9 & Acute kidney injury \\
N17.93 & AKI, stage 3 & N17.9 & Acute kidney injury \\
N17.99 & AKI, stage unspecified & N17.9 & Acute kidney injury \\
N18.0 & End-stage renal disease (legacy) & N18.6, N18.5 & Chronic kidney disease / Dialysis dependence \\
N18.2 & CKD stage 2 & N18.2 & Chronic kidney disease \\
N18.3 & CKD stage 3 & N18.3, N18.30, N18.31, N18.32 & Chronic kidney disease \\
N18.4 & CKD stage 4 & N18.4 & Chronic kidney disease \\
N18.5 & CKD stage 5 (incl. dialysis) & N18.5, N18.6 & Chronic kidney disease \\
N18.9 & CKD, unspecified & N18.9 & Chronic kidney disease \\
N18.82 & Other CKD (non-standard) & N18.9 & Chronic kidney disease \\
N18.89 & Other CKD, unspecified stage & N18.9 & Chronic kidney disease \\
R73.9 & Hyperglycaemia, unspecified & R73.9 & Hyperglycaemia \\
R99 & Ill-defined cause of mortality & R99 & Death (vital status) \\
S32.01 & Fracture, L1 vertebra & S32.01 & Vertebral fracture \\
S42.3 & Fracture, humerus shaft & S42.3 & Upper-extremity fracture \\
S42.20 & Fracture, proximal humerus, unspecified & S42.20 & Upper-extremity fracture \\
S42.21 & Fracture, surgical neck of humerus & S42.21 & Upper-extremity fracture \\
S52.50 & Distal radius fracture, unspecified & S52.50 & Upper-extremity fracture \\
S52.51 & Distal radius fracture (Colles) & S52.5 & Upper-extremity fracture \\
S52.59 & Other distal radius fractures & S52.5 & Upper-extremity fracture \\
S72.00 & Femoral neck fracture, unspecified & S72.0 & Hip/femur fracture \\
S72.10 & Pertrochanteric fracture & S72.1 & Hip/femur fracture \\
S82.6 & Fracture, lateral malleolus & S82.6 & Lower-extremity fracture \\
S82.82 & Fracture, other lower leg & S82.89 & Lower-extremity fracture \\
Z99.1 & Long-term respirator dependence & Z99.1, Z99.11 & Respirator dependence (resp. failure proxy) \\
Z99.2 & Dialysis dependence & Z99.2 & Dialysis dependence \\
All-cause mortality & All-cause mortality & (hospital\_expire\_flag / dod) & Death (vital status) \\
\end{longtable}

\begin{figure*}
    \centering
    \includegraphics[width=\textwidth]{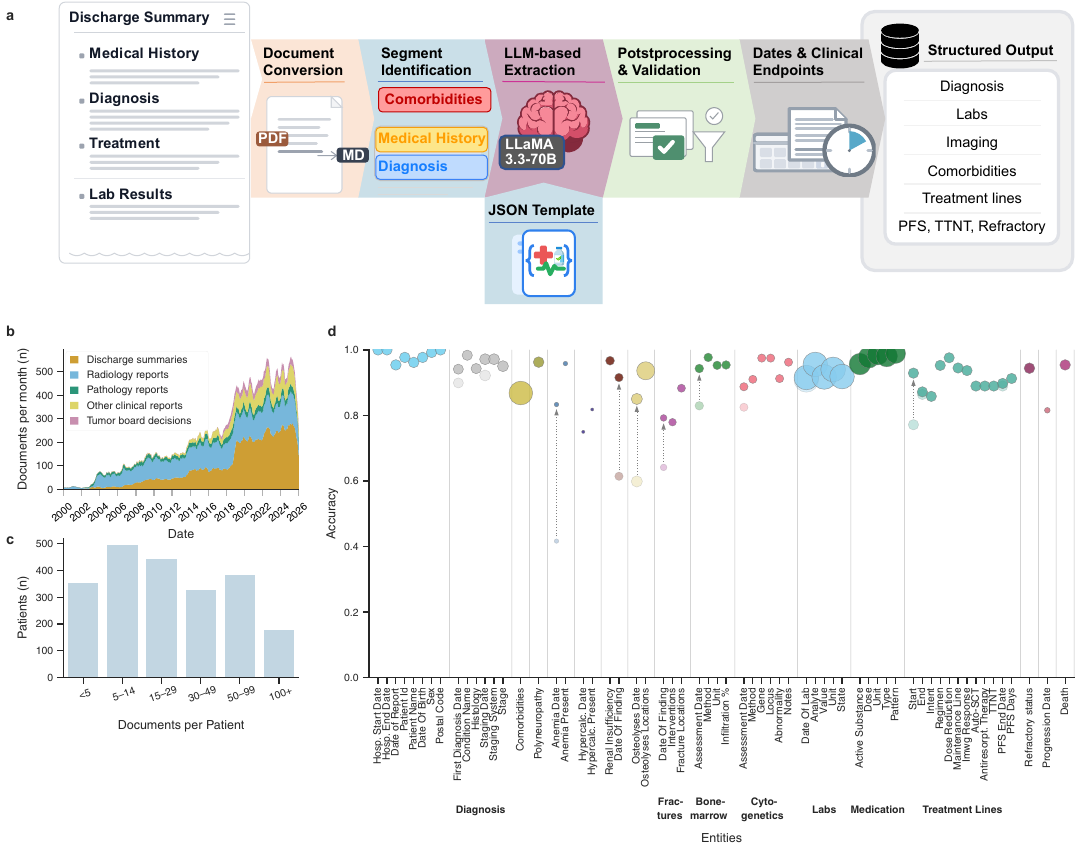}
    \caption{\textbf{Automated extraction of structured clinical data from unstructured medical documents.}
\textbf{a},~Overview of the information extraction pipeline.
Clinical documents are converted from PDF to markdown,
segmented into sections (comorbidities, medical history,
diagnosis, treatment, laboratory results), and processed
through LLaMA 3.3-70B using domain-specific JSON templates.
Extracted entities undergo postprocessing, validation, and
temporal alignment to produce structured representations
spanning diagnosis, laboratory measurements, imaging,
comorbidities, treatment lines, and clinical endpoints.
\textbf{b},~Volume of clinical documents processed over time, stratified by document type ($n = 2{,}328$ patients, 2000--2026). Discharge summaries constitute the majority of records, with increasing document volumes over the observation period.
\textbf{c},~Distribution of document counts per patient, illustrating the heterogeneity in clinical documentation density across the cohort.
\textbf{d},~Field-level extraction accuracy (exact match) across 60 clinically relevant variables grouped by domain. Each point represents one extracted variable. Point size is proportional to the number of annotated instances ($N$). For date variables, arrows extending upward from the opaque data points to semi-transparent points indicate the change from strict to relaxed date matching, which permits a broader temporal window. Accuracy exceeds 0.90 for the majority of variables, with lower performance observed for date fields with few occurences (e.g., anemia, hypercalcemia, fractures). Detailed per-variable results including relaxed accuracy metrics are provided in Table~\ref{tab:entity_accuracy_grouped_N}.}
    \label{fig:structuring_overview}
\end{figure*}

\clearpage
\section{Supplementary Tables}

\begingroup
\small
\setlength{\LTleft}{\fill}
\setlength{\LTright}{\fill}
\begin{longtable}{@{} l l c c c @{}}
\caption{
Field-level extraction accuracy across clinically relevant domains.
Accuracy denotes exact-match agreement between automated extraction and expert annotation.
\emph{Accuracy (relaxed)} reports accuracy when allowing a broader temporal window for date matching.
$N$ indicates the total number of annotated instances evaluated per variable.
}
\label{tab:entity_accuracy_grouped_N} \\
\toprule
\textbf{Clinical domain} & \textbf{Extracted variable} & \textbf{Accuracy} & \textbf{Accuracy (relaxed)} & \textbf{N} \\
\hiderowcolors
\midrule
\endfirsthead

\toprule
\textbf{Clinical domain} & \textbf{Extracted variable} & \textbf{Accuracy} & \textbf{Accuracy (relaxed)} & \textbf{N} \\
\midrule
\endhead

\midrule
\multicolumn{5}{r@{}}{\emph{Continued on next page}} \\
\endfoot

\bottomrule
\endlastfoot

\multirow{4}{*}{Patient demographics}
 & Patient name                         & 0.962 & --    & 131 \\
 & Date of birth                        & 0.977 & 0.977 & 131 \\
 & Sex                                  & 0.992 & --    & 131 \\
 & Postal code                          & 1.000 & --    & 131 \\
\midrule
\multirow{4}{*}{Hospitalization}
 & Document generation date             & 0.954 & 0.954 & 131 \\
 & Hospital admission date              & 1.000 & 1.000 & 131 \\
 & Hospital discharge date              & 1.000 & 1.000 & 129 \\
 & Patient identifier                   & 0.977 & --    & 131 \\
\end{longtable}
\endgroup

\begingroup
\small
\setlength{\LTleft}{\fill}
\setlength{\LTright}{\fill}
\begin{longtable}{@{} p{4.0cm} p{5.7cm} p{2.0cm} p{2.0cm} @{}}
\caption{
Field-level extraction accuracy across clinically relevant domains on the external dataset (document-level evaluation).
Accuracy denotes exact-match agreement between automated extraction and expert annotation.
$N$ indicates the total number of annotated instances evaluated per variable.
}
\label{tab:entity_external_accuracy_doc_level} \\
\toprule
\textbf{Clinical domain} & \textbf{Extracted variable} & \textbf{Accuracy} & \textbf{N} \\
\midrule
\endfirsthead

\toprule
\textbf{Clinical domain} & \textbf{Extracted variable} & \textbf{Accuracy} & \textbf{N} \\
\midrule
\endhead

\midrule
\multicolumn{4}{r@{}}{\emph{Continued on next page}} \\
\endfoot

\bottomrule
\endlastfoot

\multirow{4}{*}{Patient demographics}
 & Patient name                             & 1.000 & 13 \\
 & Date of birth                            & 1.000 & 13 \\
 & Sex                                      & 1.000 & 13 \\
 & Postal code                              & 0.846 & 13 \\
\midrule
\multirow{4}{*}{Hospitalization}
 & Document generation date                 & 0.769 & 13 \\
 & Hospital admission date                  & 0.846 & 13 \\
 & Hospital discharge date                  & 0.846 & 13 \\
 & Patient identifier                       & 1.000 & 9 \\
\end{longtable}
\endgroup

\rowcolors{2}{white}{gray!8}
\setlength{\LTleft}{\fill}
\setlength{\LTright}{\fill}
\begin{longtable}{@{} p{3.5cm} r r r r r r r
  @{\hspace{6pt}\vrule width 0.4pt\hspace{6pt}} r r @{}}
\caption{\textbf{Predictive performance and prediction horizon across diagnostic endpoints ordered by prevalence.}
Model performance is reported using median area under the receiver operating characteristic curve (AUROC) with 95\% confidence intervals (CI) from repeated resampling for the Transformer. For XGBoost and logistic regression, only the median AUROC is reported. Median average precision (AP) is additionally reported for the Transformer. The prediction horizon denotes the median time in days from the end of the laboratory input window to the diagnosis event, with interquartile range (IQR). Evaluated on laboratory measurements sampled from randomly positioned 10--30\,day intervals prior to diagnosis. Held-out test set comprising 1{,}579 sequences. Rows are ordered by decreasing number of positive cases ($N$). Endpoints with no positive cases in the test set are marked ``--''.}
\label{tab:performance_comparison}\\

\hiderowcolors
\toprule
& \multicolumn{4}{c}{\textbf{Transformer}} & \textbf{XGBoost} & \textbf{Log.\ Reg.} & & \multicolumn{2}{c}{\textbf{Horizon}} \\
\cmidrule(lr){2-5} \cmidrule(lr){6-6} \cmidrule(lr){7-7} \cmidrule(l){9-10}
& \multicolumn{3}{c}{AUROC} & \multicolumn{1}{c}{AP} & \multicolumn{1}{c}{AUROC} & \multicolumn{1}{c}{AUROC} & & \multicolumn{1}{c}{Median} & \multicolumn{1}{c}{IQR} \\
\cmidrule(lr){2-4} \cmidrule(lr){5-5} \cmidrule(lr){6-6} \cmidrule(lr){7-7} \cmidrule(l){9-10}
Endpoint & Median & CI$_{95}$ Low & CI$_{95}$ High & Median & Median & Median & $N$ & (days) & (days) \\
\midrule
\endfirsthead

\hiderowcolors
\toprule
& \multicolumn{4}{c}{\textbf{Transformer}} & \textbf{XGBoost} & \textbf{Log.\ Reg.} & & \multicolumn{2}{c}{\textbf{Horizon}} \\
\cmidrule(lr){2-5} \cmidrule(lr){6-6} \cmidrule(lr){7-7} \cmidrule(l){9-10}
& \multicolumn{3}{c}{AUROC} & \multicolumn{1}{c}{AP} & \multicolumn{1}{c}{AUROC} & \multicolumn{1}{c}{AUROC} & & \multicolumn{1}{c}{Median} & \multicolumn{1}{c}{IQR} \\
\cmidrule(lr){2-4} \cmidrule(lr){5-5} \cmidrule(lr){6-6} \cmidrule(lr){7-7} \cmidrule(l){9-10}
Endpoint & Median & CI$_{95}$ Low & CI$_{95}$ High & Median & Median & Median & $N$ & (days) & (days) \\
\midrule
\endhead

\midrule
\multicolumn{10}{r}{\textit{Continued on next page}}
\endfoot

\bottomrule
\endlastfoot
\showrowcolors

All-cause mortality \newline {\footnotesize All-cause mortality} & 0.697 & 0.653 & 0.737 & 0.213 & 0.656 & 0.660 & 187 & 346 & 195--531.5 \\
B96.2 \newline {\footnotesize Escherichia coli} & 0.661 & 0.616 & 0.703 & 0.256 & 0.628 & 0.633 & 175 & 210 & 69--450.5 \\
I10.00 \newline {\footnotesize Essential hypertension, benign} & 0.722 & 0.680 & 0.763 & 0.291 & 0.680 & 0.700 & 159 & 112 & 24--283 \\
C78.6 \newline {\footnotesize Sec.\ malig.\ neoplasm, peritoneum} & 0.888 & 0.862 & 0.913 & 0.471 & 0.882 & 0.870 & 117 & 97 & 20--255 \\
B95.2 \newline {\footnotesize Streptococcus grp D / Enterococcus} & 0.660 & 0.609 & 0.715 & 0.124 & 0.637 & 0.689 & 110 & 230.5 & 84--426 \\
I10.90 \newline {\footnotesize Essential hypertension, unspec.} & 0.768 & 0.723 & 0.807 & 0.176 & 0.659 & 0.735 & 97 & 139 & 43--330 \\
E03.8 \newline {\footnotesize Other hypothyroidism} & 0.694 & 0.640 & 0.749 & 0.105 & 0.674 & 0.686 & 86 & 210 & 42.2--365 \\
M89.50 \newline {\footnotesize Osteolysis, multiple sites} & 0.847 & 0.813 & 0.876 & 0.238 & 0.857 & 0.870 & 85 & 192 & 96--350 \\
\end{longtable}
\rowcolors{1}{white}{white}
 

\rowcolors{2}{white}{white}
\setlength{\LTleft}{\fill}
\setlength{\LTright}{\fill}
\definecolor{grouprow}{gray}{0.92}
\begin{longtable}{@{} p{4.0cm} r r r r r r r
  @{\hspace{6pt}\vrule width 0.4pt\hspace{6pt}} r r @{}}
\caption{\textbf{Predictive performance and prediction horizon across grouped diagnostic endpoints.}
Performance is summarised at two levels of aggregation: the bold rows give pooled metrics over all endpoints in a clinical category, the indented rows show the corresponding three-character ICD-10 sub-categories. Transformer AUROC is reported with 95\% confidence intervals (CI) from repeated resampling; XGBoost and logistic regression AUROC are reported as medians; median average precision (AP) is additionally reported for the Transformer. Prediction horizon is the median time in days from the end of the laboratory input window to the diagnosis event, with interquartile range (IQR). Held-out test set comprising 1{,}579 sequences; $N$ is the number of positive cases in the test set.}
\label{tab:performance_comparison_grouped}\\
\toprule
& \multicolumn{4}{c}{\textbf{Transformer}} & \textbf{XGBoost} & \textbf{Log.\ Reg.} & & \multicolumn{2}{c}{\textbf{Horizon}} \\
\cmidrule(lr){2-5} \cmidrule(lr){6-6} \cmidrule(lr){7-7} \cmidrule(l){9-10}
& \multicolumn{3}{c}{AUROC} & \multicolumn{1}{c}{AP} & \multicolumn{1}{c}{AUROC} & \multicolumn{1}{c}{AUROC} & & \multicolumn{1}{c}{Median} & \multicolumn{1}{c}{IQR} \\
\cmidrule(lr){2-4} \cmidrule(lr){5-5} \cmidrule(lr){6-6} \cmidrule(lr){7-7} \cmidrule(l){9-10}
Endpoint & Median & CI$_{95}$ Low & CI$_{95}$ High & Median & Median & Median & $N$ & (days) & (days) \\
\midrule
\endfirsthead

\toprule
& \multicolumn{4}{c}{\textbf{Transformer}} & \textbf{XGBoost} & \textbf{Log.\ Reg.} & & \multicolumn{2}{c}{\textbf{Horizon}} \\
\cmidrule(lr){2-5} \cmidrule(lr){6-6} \cmidrule(lr){7-7} \cmidrule(l){9-10}
& \multicolumn{3}{c}{AUROC} & \multicolumn{1}{c}{AP} & \multicolumn{1}{c}{AUROC} & \multicolumn{1}{c}{AUROC} & & \multicolumn{1}{c}{Median} & \multicolumn{1}{c}{IQR} \\
\cmidrule(lr){2-4} \cmidrule(lr){5-5} \cmidrule(lr){6-6} \cmidrule(lr){7-7} \cmidrule(l){9-10}
Endpoint & Median & CI$_{95}$ Low & CI$_{95}$ High & Median & Median & Median & $N$ & (days) & (days) \\
\midrule
\endhead
\midrule
\multicolumn{10}{r}{\textit{Continued on next page}}
\endfoot

\bottomrule
\endlastfoot
\rowcolor{grouprow} \textbf{Cardiovascular diseases} & \textbf{0.682} & \textbf{0.651} & \textbf{0.709} & \textbf{0.480} & \textbf{0.589} & \textbf{0.629} & \textbf{470} & \textbf{113} & \textbf{28.2--356} \\
\quad I10 \newline {\footnotesize Essential hypertension} & 0.917 & 0.862 & 0.967 & 0.004 & 0.786 & 0.825 & 263 & 92 & 25--279.5 \\
\quad I48 \newline {\footnotesize Atrial fibrillation/flutter} & 0.741 & 0.695 & 0.788 & 0.188 & 0.671 & 0.752 & 110 & 187 & 52.2--411.2 \\
\quad I80 \newline {\footnotesize Phlebitis/thrombophlebitis} & 0.601 & 0.544 & 0.654 & 0.088 & 0.582 & 0.638 & 99 & 177 & 72.5--360 \\
\quad I50 \newline {\footnotesize Heart failure} & 0.683 & 0.616 & 0.749 & 0.156 & 0.594 & 0.631 & 76 & 175 & 70--482.5 \\
\quad I25 \newline {\footnotesize Chronic ischaemic heart disease} & 0.705 & 0.640 & 0.769 & 0.108 & 0.561 & 0.673 & 70 & 147 & 19.2--362.8 \\
\quad I11 \newline {\footnotesize Hypertensive heart disease} & 0.815 & 0.700 & 0.902 & 0.119 & 0.625 & 0.727 & 22 & 191.5 & 37.8--547 \\
\quad I12 \newline {\footnotesize Hypertensive chronic kidney disease} & 0.788 & 0.670 & 0.886 & 0.030 & 0.685 & 0.771 & 12 & 445 & 279.5--547.5 \\
\end{longtable}

\setlength{\LTleft}{\fill}
\setlength{\LTright}{\fill}
\rowcolors{2}{white}{gray!8}
\begin{longtable}{@{} p{3.5cm} r r r r r
  @{\hspace{6pt}\vrule width 0.4pt\hspace{6pt}} r r r r r @{}}
\caption{\textbf{External validation performance per ICD-10-GM endpoint on MIMIC-IV and MMRF CoMMpass.}
For each of the 162 target endpoints we report area under the receiver-operating-characteristic curve (AUROC) with bootstrap 95\% confidence interval, average precision (AP), and the number of positive samples ($N$) on the full MIMIC-IV and the MMRF CoMMpass IA24 datasets without finetuning. AUROC and AP values are bootstrap medians. $N$ is computed as $\lfloor\pi \times N_{\mathrm{eval}}\rceil$ where $\pi$ is the median positive-class prevalence and $N_{\mathrm{eval}}^{\mathrm{MIMIC}}=8150$, $N_{\mathrm{eval}}^{\mathrm{MMRF}}=7347$ samples. Endpoints with zero positive cases in a given cohort are reported as ``--''. Rows are ordered by the larger of the two cohort prevalences, descending.}

\label{tab:performance_comparison_extern} \\
\hiderowcolors
\toprule
 & \multicolumn{5}{c}{\textbf{MIMIC-IV}} & \multicolumn{5}{c}{\textbf{MMRF CoMMpass}} \\
\cmidrule(lr){2-6}\cmidrule(l){7-11}
 & \multicolumn{3}{c}{AUROC} & \multicolumn{1}{c}{AP} & $N$ & \multicolumn{3}{c}{AUROC} & \multicolumn{1}{c}{AP} & $N$ \\
\cmidrule(lr){2-4}\cmidrule(lr){5-5}\cmidrule(lr){6-6}\cmidrule(lr){7-9}\cmidrule(lr){10-10}\cmidrule(l){11-11}
Endpoint & Median & CI$_{95}$ Low & CI$_{95}$ High & Median & & Median & CI$_{95}$ Low & CI$_{95}$ High & Median & \\
\midrule
\endfirsthead
\hiderowcolors
\toprule
 & \multicolumn{5}{c}{\textbf{MIMIC-IV}} & \multicolumn{5}{c}{\textbf{MMRF CoMMpass}} \\
\cmidrule(lr){2-6}\cmidrule(l){7-11}
 & \multicolumn{3}{c}{AUROC} & \multicolumn{1}{c}{AP} & $N$ & \multicolumn{3}{c}{AUROC} & \multicolumn{1}{c}{AP} & $N$ \\
\cmidrule(lr){2-4}\cmidrule(lr){5-5}\cmidrule(lr){6-6}\cmidrule(lr){7-9}\cmidrule(lr){10-10}\cmidrule(l){11-11}
Endpoint & Median & CI$_{95}$ Low & CI$_{95}$ High & Median & & Median & CI$_{95}$ Low & CI$_{95}$ High & Median & \\
\midrule
\endhead
\midrule
\multicolumn{11}{r}{\textit{Continued on next page}} \\
\endfoot
\bottomrule
\endlastfoot

\showrowcolors
All-cause mortality \newline {\footnotesize All-cause mortality} & 0.635 & 0.622 & 0.648 & 0.432 & 2356 & 0.691 & 0.677 & 0.704 & 0.324 & 1582 \\
M89.59 \newline {\footnotesize Osteolysis, unspec.\ site} & -- & -- & -- & -- & 0 & 0.553 & 0.514 & 0.587 & 0.227 & 1378 \\
M89.55 \newline {\footnotesize Osteolysis, femur} & -- & -- & -- & -- & 0 & 0.595 & 0.560 & 0.631 & 0.243 & 1378 \\
M89.58 \newline {\footnotesize Osteolysis, other site} & 0.380 & 0.229 & 0.586 & 0.001 & 8 & 0.571 & 0.534 & 0.606 & 0.237 & 1378 \\
M89.50 \newline {\footnotesize Osteolysis, multi-site/unspec.} & -- & -- & -- & -- & 0 & 0.582 & 0.546 & 0.617 & 0.231 & 1378 \\
I10.91 \newline {\footnotesize Essential HTN, with crisis} & 0.516 & 0.498 & 0.536 & 0.140 & 1068 & 0.741 & 0.700 & 0.779 & 0.102 & 254 \\
I10.01 \newline {\footnotesize Malignant essential HTN} & 0.528 & 0.510 & 0.549 & 0.147 & 1068 & 0.675 & 0.631 & 0.721 & 0.063 & 254 \\
I10 \newline {\footnotesize Essential primary hypertension} & 0.472 & 0.453 & 0.492 & 0.138 & 1068 & 0.551 & 0.502 & 0.597 & 0.038 & 254 \\
\end{longtable}

\rowcolors{2}{white}{white}
\setlength{\LTleft}{\fill}
\setlength{\LTright}{\fill}
\definecolor{grouprow}{gray}{0.92}
\begin{longtable}{@{} p{4.0cm} r r r r r
  @{\hspace{6pt}\vrule width 0.4pt\hspace{6pt}} r r r r r @{}}
\caption{\textbf{External validation performance across grouped diagnostic endpoints on MIMIC-IV and MMRF CoMMpass.}
Performance is summarised at two levels of aggregation: the bold rows give pooled metrics over all endpoints in a clinical category, the indented rows show the corresponding three-character ICD-10 sub-categories. AUROC and average precision (AP) are bootstrap medians; CI$_{95}$ bounds are the 2.5\textsuperscript{th} and 97.5\textsuperscript{th} percentiles. The number of positive samples ($N$) is computed as $\lfloor\pi \times N_{\mathrm{eval}}\rceil$ where $\pi$ is the median positive-class prevalence and $N_{\mathrm{eval}}^{\mathrm{MIMIC}}=8150$, $N_{\mathrm{eval}}^{\mathrm{MMRF}}=7347$ samples. Endpoints with zero positive cases in a given cohort are reported as ``--''. Rows are ordered by the larger of the two cohort prevalences, descending.}

\label{tab:performance_comparison_extern_grouped}\\

\toprule
 & \multicolumn{5}{c}{\textbf{MIMIC-IV}} & \multicolumn{5}{c}{\textbf{MMRF CoMMpass}} \\
\cmidrule(lr){2-6}\cmidrule(l){7-11}
 & \multicolumn{3}{c}{AUROC} & \multicolumn{1}{c}{AP} & $N$ & \multicolumn{3}{c}{AUROC} & \multicolumn{1}{c}{AP} & $N$ \\
\cmidrule(lr){2-4}\cmidrule(lr){5-5}\cmidrule(lr){6-6}\cmidrule(lr){7-9}\cmidrule(lr){10-10}\cmidrule(l){11-11}
Endpoint & Median & CI$_{95}$ Low & CI$_{95}$ High & Median & & Median & CI$_{95}$ Low & CI$_{95}$ High & Median & \\
\midrule
\endfirsthead
\toprule
 & \multicolumn{5}{c}{\textbf{MIMIC-IV}} & \multicolumn{5}{c}{\textbf{MMRF CoMMpass}} \\
\cmidrule(lr){2-6}\cmidrule(l){7-11}
 & \multicolumn{3}{c}{AUROC} & \multicolumn{1}{c}{AP} & $N$ & \multicolumn{3}{c}{AUROC} & \multicolumn{1}{c}{AP} & $N$ \\
\cmidrule(lr){2-4}\cmidrule(lr){5-5}\cmidrule(lr){6-6}\cmidrule(lr){7-9}\cmidrule(lr){10-10}\cmidrule(l){11-11}
Endpoint & Median & CI$_{95}$ Low & CI$_{95}$ High & Median & & Median & CI$_{95}$ Low & CI$_{95}$ High & Median & \\
\midrule
\endhead

\midrule
\multicolumn{11}{r}{\textit{Continued on next page}}
\endfoot

\bottomrule
\endlastfoot
\rowcolor{grouprow} \textbf{All-cause mortality} & \textbf{0.635} & \textbf{0.622} & \textbf{0.648} & \textbf{0.432} & \textbf{2356} & \textbf{0.691} & \textbf{0.677} & \textbf{0.704} & \textbf{0.324} & \textbf{1582} \\
\midrule
\rowcolor{grouprow} \textbf{Cardiovascular diseases} & \textbf{0.593} & \textbf{0.578} & \textbf{0.608} & \textbf{0.277} & \textbf{1645} & \textbf{0.669} & \textbf{0.648} & \textbf{0.690} & \textbf{0.141} & \textbf{674} \\
\quad I10 \newline {\footnotesize Essential hypertension} & 0.472 & 0.453 & 0.492 & 0.138 & 1068 & 0.551 & 0.502 & 0.597 & 0.038 & 254 \\
\quad I48 \newline {\footnotesize Atrial fibrillation/flutter} & 0.643 & 0.620 & 0.666 & 0.091 & 411 & 0.682 & 0.616 & 0.737 & 0.021 & 88 \\
\quad I80 \newline {\footnotesize Phlebitis/thrombophlebitis} & 0.492 & 0.454 & 0.531 & 0.021 & 161 & 0.549 & 0.465 & 0.616 & 0.009 & 52 \\
\quad I50 \newline {\footnotesize Heart failure} & 0.652 & 0.629 & 0.675 & 0.090 & 423 & 0.590 & 0.497 & 0.673 & 0.025 & 61 \\
\quad I25 \newline {\footnotesize Chronic ischaemic heart disease} & 0.686 & 0.664 & 0.713 & 0.074 & 321 & 0.591 & 0.501 & 0.670 & 0.008 & 54 \\
\quad I11 \newline {\footnotesize Hypertensive heart disease} & 0.621 & 0.585 & 0.657 & 0.033 & 169 & 0.619 & 0.577 & 0.659 & 0.054 & 289 \\
\end{longtable}

\section{Supplementary Results}

\subsection{Information extraction from unstructured clinical documents}

Extraction accuracy was evaluated by exact match against expert annotations across 60 clinically relevant variables grouped into 15 clinical domains (Extended Data Fig.~\ref{fig:structuring_overview}d, Table~\ref{tab:entity_accuracy_grouped_N}). Accuracy exceeded 0.90 for most variables, including laboratory analyte values (accuracy $>$ 0.95) and medication active substances, doses, and units (accuracy $>$ 0.90). Core diagnostic fields, treatment line variables (0.85 to 0.95), refractory status, and cytogenetic abnormality fields also exceeded 0.90. Date fields showed greater variability, with strict exact-match accuracy ranging from 0.40 to 0.95 depending on domain. Under relaxed date matching permitting a one-month temporal window, accuracy improved by up to 45 percentage points for affected variables (Table~\ref{tab:entity_accuracy_grouped_N}).

External validation on 14 documents from four independent institutions confirmed transferability of extraction performance (Table~\ref{tab:entity_external_accuracy_doc_level}). Demographic fields, diagnosis, bone marrow assessment, and death date achieved perfect or near-perfect accuracy. Medication extraction showed slightly lower accuracy on external documents (active substance 0.895, dose 0.870) compared with internal documents, consistent with greater variability in document formatting across institutions. Laboratory measurement accuracy remained above 0.85 across all external fields.

\subsection{Comparison with non-sequential baselines}

Per-endpoint and per-diagnosis performance comparisons between the transformer, XGBoost, and logistic regression are provided in Supplementary Tables~\ref{tab:performance_comparison},~\ref{tab:performance_comparison_grouped}. Baseline methodology is described in Methods.

\subsection{Model calibration}
Model calibration was assessed using Platt scaling fitted independently per diagnosis group on the validation set. Predictions were binned into ten equally large quantiles (Figure~\ref{fig:model_comparison}b).
After scaling, all three models were well calibrated, and their quantile bins overlapped closely and spanned a comparable range of predicted probabilities for each endpoint. The majority of predictions fell within the 0.0--0.2 range, where the binned predictions of all three models tracked the calibration diagonal closely. For the higher-prevalence endpoints (all-cause mortality, metastatic disease, cardiovascular disease, kidney disease and bacterial infections), predictions extended past 0.3 predicted probability, reaching approximately 0.5--0.6 for metastatic and cardiovascular disease, and remained close to the diagonal, with the highest-probability bins tending to fall slightly below it, indicating mild over-prediction in the sparsely populated upper range.
For the low-prevalence endpoints (myelodysplastic syndromes, type~2 diabetes and fungal infections), predictions were confined to the lower-frequency region, limiting the assessable calibration range but showing close agreement with the diagonal within that region. Notably, the transformer's discrimination advantage over the non-sequential baselines (Figure~\ref{fig:model_comparison}a) did not translate into a corresponding calibration difference: after per-group Platt scaling, the XGBoost and logistic regression bins coincided with those of the transformer at equivalent predicted probabilities across all eight diagnosis groups.

\subsection{Subgroup stratification and ablation}
The transformer was evaluated under stratified and ablated conditions (Figure~\ref{fig:subgroup_ablation}); all values are median AUROC across all diagnoses. Performance was broadly preserved across age strata (Figure~\ref{fig:subgroup_ablation}b): median AUROC 0.70--0.71 across the three older strata (55--64: 0.708; 65--74: 0.710; 75+: 0.703) and 0.762 in the youngest (18--54). Female patients achieved higher median AUROC than male patients (0.721 vs.\ 0.678), with overlapping 95\% bootstrap confidence intervals; the gap likely reflects cohort composition, as the ovarian cancer subcohort contributing approximately half the training data is exclusively female.

Ablating demographic inputs or the imputation module had divergent effects on the internal and external cohorts (Figure~\ref{fig:subgroup_ablation}a). On the internal test set, the full model (sex, age, and laboratory imputation) achieved a median AUROC of 0.740; removing sex ($-$0.004), age ($+$0.004), both demographic variables ($+$0.002), or laboratory imputation ($+$0.001) each changed median AUROC by less than 0.01, indicating that internal discrimination derives almost entirely from the laboratory measurements. Using the full model while disabling the ensemble reduced median AUROC by 0.037. On the external cohorts the same ablations produced larger degradations. On MIMIC-IV (full model 0.642), removing sex, age, or both lowered AUROC by 0.013 to 0.023, with the largest drop for removing both demographic variables ($-$0.023); omitting imputation reduced AUROC by 0.021. On MMRF CoMMpass (full model 0.616), removing sex produced the largest single drop ($-$0.029), followed by age ($-$0.024), both demographics ($-$0.019), and no imputation ($-$0.015). The contrast indicates that demographic features and the imputation module, while largely redundant with the laboratory signal internally, contribute materially to cross-cohort generalisation, consistent with the masking analysis: on MMRF the derived glomerular filtration estimates are unavailable, so the reconstruction path that internally compensates for missing demographic information is absent.

All values reported here are medians across individual ICD codes rather than the grouped endpoints used in the main text. At the level of individual codes, differences between datasets partly reflect inconsistent definitions and coding practices across cohorts rather than genuine differences in model performance. Grouping subcodes into broader categories reconciles these definitional discrepancies, so the cross-dataset differences are larger at the per-code level than for the grouped endpoints.

\subsection{Per-diagnosis performance within grouped endpoints}

Performance varied across individual ICD-10 codes within several grouped endpoints (Supplementary Table~\ref{tab:performance_comparison}). Within the metastatic group, C78 (secondary malignant neoplasms of respiratory and digestive organs; AUROC 0.82, 2.9-fold enrichment) was the strongest sub-category, with peritoneal metastases (C78.6; AUROC 0.89, 5.1-fold enrichment) the strongest individual contributor, whereas other-site metastases (C79; AUROC 0.61, 1.5-fold) showed only modest enrichment. Kidney disease discrimination was driven primarily by chronic kidney disease (N18; AUROC 0.89, 5.2-fold enrichment) rather than acute kidney failure (N17; AUROC 0.66, 1.8-fold). 
End-stage renal disease (N18.5; AUROC 0.90) and dialysis dependence (Z99.2; AUROC 0.81) showed the highest enrichment of any code present in at least 1\% of test cases (both approximately 25-fold), although their confidence intervals were wide. Sepsis (A41; AUROC 0.74, 3.1-fold enrichment) showed stronger enrichment than the more prevalent gram-positive (B95; AUROC 0.64, 1.6-fold) and gram-negative (B96; AUROC 0.66, 1.9-fold) organism codes that dominated the bacterial infections group. Among individual cardiovascular endpoints, atrial fibrillation (I48; AUROC 0.74, 2.8-fold) and chronic ischaemic heart disease (I25; AUROC 0.71, 2.5-fold) exceeded the group-level AUROC of 0.68, while heart failure (I50; AUROC 0.68, 3.3-fold) showed the strongest enrichment at comparable discrimination. Among rare diagnoses with fewer than 10 positive cases in the test set, confidence intervals widened substantially. Full per-diagnosis results including confidence intervals for all 162 endpoints are provided in Supplementary Table~\ref{tab:performance_comparison}.

\section{Supplementary Methods}

\subsection{Clinically defined target entities}
\label{sec:clin_targets}

The following clinical entities, identified by oncology and radiology experts, informed the design of extraction templates, validation rules, and the evaluation framework (Supplementary Table~\ref{tab:entity_accuracy_grouped_N}).

\subsubsection{Patient demographics}

Six demographic variables were extracted from each document, namely patient identifier, full name, date of birth, sex, postal code of residence, and document date. These fields served as identifiers for longitudinal record linkage and may be used as enriching input features (age, sex) for downstream prediction models.

\subsubsection{Hospitalisation}

For each inpatient episode, hospital admission and discharge dates were extracted. These dates defined encounter boundaries and established chronological ordering of clinical events. Hospitalisation dates served as temporal anchors when the exact date of a diagnosis was not documented.

\subsubsection{Comorbidities}

Pre-existing and incident comorbid conditions were extracted as diagnosis name and date of diagnosis pairs. Diagnoses were mapped to ICD-10 codes where available in the source text.

\subsubsection{Primary diagnosis and staging}

The primary oncological diagnosis was captured as a diagnosis name, date of first diagnosis, and histological subtype (e.g., IgG-kappa, IgA-lambda for myeloma). Disease staging was extracted separately for each staging assessment documented in the clinical record, comprising the staging system used (e.g., International Staging System, Revised ISS, Durie-Salmon), the assigned tumour stage, and the date of the staging assessment.

\subsubsection{Cytogenetic assessment}

Cytogenetic findings were extracted per reported abnormality, including the assessment date, the method used (fluorescence in situ hybridisation or conventional karyotyping), the affected gene (e.g., TP53, FGFR3), the genomic locus (e.g., 17p13, 4p16), and the type of detected abnormality (deletion, translocation, gain). Domain-specific high-risk markers including del(17p), t(4;14), t(14;16), gain(1q), and del(1p) were flagged for downstream risk stratification.

\subsubsection{Bone marrow assessment}

Each bone marrow evaluation was represented by the assessment date, the method used (aspiration, biopsy, or flow cytometry), the measured infiltration percentage, and the corresponding measurement unit.

\subsubsection{Medication}

Pharmacological treatments were extracted at the level of individual prescriptions, comprising the active substance name, dose, unit, administration schedule type (e.g., continuous, intermittent, as needed), and administration schedule pattern. The schedule pattern encoded either cycle-based regimens (e.g., days 1, 8, 15 of a 28-day cycle) for chemotherapy, or daily dosing distributions using the standard morning-noon-evening-night notation (e.g., 1-0-0 for once daily in the morning, 0-1-0-1 for noon and night administration) for supportive and chronic medications.

\subsubsection{Laboratory measurements}

Each laboratory result was extracted as a tuple of assessment date, analyte name, measured value, unit, and result status. The result status field indicated whether the measurement represented a standard validated result, point-of-care or preliminary measurement, or was flagged as invalid (e.g., haemolysed sample, insufficient material). Point-of-care and invalid results were retained in the extracted dataset but flagged for optional exclusion in downstream analyses.

\subsubsection{Treatment lines}

The course of disease was reconstructed as a sequence of treatment lines, each comprising treatment intent (first-line, salvage, or palliative), regimen name (e.g., VRd, KRd, Dara-VMP), treatment start and end dates, whether consolidation or maintenance therapy was administered, concurrent antiresorptive therapy (e.g., zoledronic acid, denosumab), dose reduction (yes/no), administration of high-dose therapy with autologous stem cell transplantation (yes/no), best response according to the International Myeloma Working Group uniform response criteria, progression-free survival (event occurrence, event date, and duration in days), and time to next treatment.

\subsubsection{Refractory status}

Drug-specific refractoriness was defined as non-response or disease progression on therapy, or within 60 days after discontinuation of a given agent~\cite{Rajkumar2011}. Refractory status was captured for the following drug classes, including immunomodulatory agents (thalidomide, lenalidomide, pomalidomide), proteasome inhibitors (bortezomib, carfilzomib, ixazomib), monoclonal antibodies and targeted agents (anti-CD38 antibodies, elotuzumab, belantamab mafodotin, selinexor), and historical therapies (interferon, melphalan, cyclophosphamide, corticosteroids). Composite refractory categories were derived, including double-class (immunomodulatory agent plus proteasome inhibitor), triple-class (immunomodulatory agent plus proteasome inhibitor plus anti-CD38 antibody), quadruple-class, penta-drug (two immunomodulatory agents, two proteasome inhibitors, and an anti-CD38 antibody), and BCMA-directed therapy refractoriness.

\subsubsection{Skeletal findings}

Bone disease was captured through three entity types, namely osteolytic lesions (date of diagnosis and anatomical locations of individual lesions), fractures (date of finding, anatomical location, and any documented interventions with intervention dates), and general skeletal assessment findings. Lesion and fracture locations were recorded as free-text anatomical descriptors and normalised to a standardised skeletal region vocabulary during postprocessing.

\subsubsection{Neurological complications}

The presence or absence of polyneuropathy was extracted as a binary variable. Polyneuropathy is a frequent treatment-emergent adverse event associated with proteasome inhibitors and immunomodulatory agents and was tracked longitudinally across documents to monitor treatment tolerability.

\subsubsection{SLiM-CRAB criteria}

Myeloma-defining events according to the SLiM-CRAB framework were extracted as composite entities, specifically bone marrow infiltration exceeding 60\% (S; assessment date, method, value), the presence of one or more osteolytic lesion on cross-sectional imaging (B; location, date), hypercalcaemia (C; date of documentation), renal insufficiency (R; date of documentation), and anaemia (A; date of documentation). The involved-to-uninvolved serum free light chain ratio (Li) was derived from extracted laboratory values where both measurements were available within the same assessment.

\subsubsection{Death}

Date of death was extracted where documented in the clinical record.

\subsection{Information extraction pipeline}

The information extraction pipeline comprised five stages operating sequentially on each clinical document (Extended Data Fig.~\ref{fig:structuring_overview}).

In the first stage, complete discharge summaries, radiology reports, and pathology reports were converted from PDF into Markdown. Formatting cues such as font type, size, text positioning, and layout metadata were used to identify and exclude non-clinical content (hospital contact information, page headers, footnotes, boilerplate disclaimers), while preserving document hierarchy including section headers, paragraph boundaries, and report metadata (Extended Data Fig.~\ref{fig:structuring_overview}a).

The resulting normalised documents were then segmented into semantically coherent units, such as diagnosis, medical history, comorbidities, treatment descriptions, laboratory results, and radiology or pathology subsections, using rule-based heuristics that relied on heading keyword dictionaries, line-prefix patterns, and section delimiters. This hierarchical segmentation ensured that the downstream extraction model received only the most relevant textual context for each target entity, reducing noise and improving extraction quality (Extended Data Fig.~\ref{fig:structuring_overview}d).

In the third stage, a locally deployed Llama 3.3-70B model performed template-driven extraction at the segment level, using a 4-bit AWQ-quantized variant (casperhansen/llama-3.3-70b-instruct-awq) to enable single-GPU deployment \cite{casperhansen_Llama3_3_70B_Instruct_AWQ_2026}. For each segment type, a dedicated template specified target fields, value constraints, and a strict JSON output schema (Listing~\ref{lst:llm_template_diagnosis}). Invoking the model at the segment level rather than on full documents limited context size and focused extraction on relevant mentions. Templates were refined with domain experts to address ambiguous cases and improve consistency across reports. In addition to structured fields, the model optionally produced short narrative justifications describing the textual evidence supporting each extracted value.

Postprocessing and validation constituted the fourth stage. The model-generated narrative justifications were compared against the corresponding structured outputs to verify consistency between extracted values and their textual evidence. Discrepancies triggered automated rejection or re-querying of the model. Field-specific rule-based checks were applied for selected entities to exclude common confounders, for example differentiating osteolytic lesions from fractures in imaging-derived bone findings. This validation framework combined narrative consistency checks with targeted domain rules to improve robustness without requiring manual curation.

Finally, temporal extraction and clinical endpoint construction resolved all date expressions into DD-MM-YYYY format using regular-expression-based date discovery and rule-based resolution of relative temporal expressions anchored to the report date (Listing~\ref{lst:regex_dates}). When the day or month was not specified, missing components were imputed using January 1 of the corresponding year. Normalised events were sorted chronologically and grouped into lines of therapy using a second LLM pass that ingested the ordered event list and output structured treatment-line representations with normalised regimens. Progression-free survival and time to next treatment were computed from progression events and treatment-line boundaries, taking the minimum of the time to the next documented progression event or initiation of the subsequent line of therapy. Refractory status was determined using the rules defined in the entity definitions.

\begin{center}
\captionsetup{type=listing}
\captionof{listing}{Example LLM extraction prompt and output schema for primary diagnosis segments}
\label{lst:llm_template_diagnosis}
\end{center}

\vspace{4pt}
\noindent\textbf{Prompt (abstracted):} ``Given the following segment from a clinical report, extract the requested fields. Output only valid JSON that conforms to the schema. If a field is not present, use \texttt{null}. For each extracted item include a short \texttt{narrative} quoting the minimal text that supports the extraction.''

\vspace{4pt}
\noindent\textbf{Output schema (example):}
\begin{verbatim}
{
  "segment_type": "primary diagnosis",
  "entities": [
    {
      "type": "Diagnosis",
      "narrative": "Free-text summary of the main oncologic diagnosis as stated in the letter.",
      "structuredDetails":
      {
        conditionName: "Multiple Myeloma" | other | -None- # if multiple myeloma is 
        not mentioned specifically as the diagnosis, name the other
        histology: e.g. "IgA Kappa" | -None-
        dateOfFirstDiagnosis: "YYYY-MM" | "YY-MM" | "MM/YY" | "YYYY" | -None- 
                            # Date of initial Myeloma diagnosis. 
                            Typically date of ED. 
                            if "Aktuell:". use date of report
        tumorStagings:
        {
            system: str | -None- # e.g. R2-ISS, R-ISS, ISS, Durie-Salmon, etc.
            stage: str | -None- # e.g. "II"
            date: "YYYY-MM" | -None- # date of the staging,
            often dateOfFirstDiagnosis 
        }
        [if applicable list other stagings mentioned]
    }
}
\end{verbatim}

\begin{center}
\captionsetup{type=listing}
\captionof{listing}{Example regular expression for extraction of dates}
\label{lst:regex_dates}
\end{center}
\begin{verbatim}
(\b\d{4}\b)                            # year only, e.g., 2020
(\b\d{1,2}[/-]\d{1,2}[/-]\d{2,4}\b)    # numeric dates: 15/07/2020 or 7-15-20
(\b(?:Jan|Feb|Mar|Apr|May|Jun|Jul|Aug|Sep|Oct|Nov|Dec)[a-z]*\.?\s+\d{1,2},?\s+\d{4}\b)
                                       # textual: July 15, 2020
\end{verbatim}

\subsubsection{Extraction performance evaluation}

Pipeline accuracy was evaluated against manually annotated reference sets comprising 131 internal discharge summaries (18,875 structured outputs) and 14 external documents from four independent institutions (3,970 structured outputs). Accuracy was computed as the proportion of extracted attribute values matching reference annotations under exact-match criteria. For temporal attributes, two evaluation modes were applied, namely strict exact-match requiring agreement to the day and a lenient window accepting dates within one calendar month of the reference. For chronic conditions, the date of hospitalisation was accepted as a valid proxy for date of diagnosis when no more precise temporal anchor was available in the source document. When the pipeline could not recover a specific date, the hospitalisation date was assigned as a pragmatic temporal anchor to preserve longitudinal usability rather than omitting the attribute entirely. These cases are scored as errors under strict exact-match evaluation and reflect a deliberate design priority of structured output completeness over annotation fidelity. Restricting evaluation to predictions with complete date information increased mean date accuracy from 77.6\% to 87.1\%, confirming that residual temporal error originated primarily from ambiguous source documentation rather than extraction failure on well-specified dates.

All annotations were reviewed by three medically trained annotators comprising two medical doctoral candidates and one physician undergoing specialist training in radiology. Annotation followed a pre-specified schema. Disagreements were resolved by discussion, with the physician annotator serving as adjudicator for clinically ambiguous cases. Full per-entity accuracy results are reported in Supplementary Tables~\ref{tab:entity_accuracy_grouped_N} and ~\ref{tab:entity_external_accuracy_doc_level}.

\subsection{Diagnosis prediction: technical details}

\subsubsection{Sources of bias in longitudinal laboratory data}

Four main sources of bias affect longitudinal laboratory data collected in routine clinical care. Testing frequency increases with disease severity and treatment complexity, resulting in overrepresentation of patients with advanced disease. Laboratory measurements are often ordered in response to clinical suspicion, introducing indication bias. Data originate from a tertiary university hospital, where case complexity is higher than in general care settings, and the cohort does not include healthy individuals. Predictions should therefore be interpreted within a clinical cancer population. Patients with longer survival contribute more observation time and thus more eligible laboratory sequences, introducing survival bias. The per-patient sequence cap described below partially mitigates this effect.

\subsubsection{Cohort construction and filtering}

Cohort construction, analyte selection, patient filtering, and dataset splits are described in Methods. The following paragraphs provide additional justification for the prediction horizon and per-patient sequence cap.

\paragraph{Choice of prediction horizon.}

A~prediction horizon of 730 days was selected to capture the treatment-associated complication burden that accumulates over one to two lines of therapy. In relapsed multiple myeloma, median progression-free survival ranges from approximately 12 to 24 months depending on treatment line and regimen~\cite{Rajkumar2022}, and median time to next treatment follows a comparable distribution. In recurrent ovarian cancer, platinum-free intervals defining sensitivity thresholds and typical re-treatment cycles similarly fall within this range~\cite{Travis1999}. A~730-day horizon therefore aligns with the temporal scale over which clinically relevant complications are expected to manifest. Shorter horizons (e.g., 90 or 180 days) would preferentially capture acute toxicities such as neutropenic infections but miss slower-onset conditions. Longer horizons would introduce excessive noise from intercurrent treatment changes and disease progression events that alter the laboratory trajectory independently of the complications being predicted.

\paragraph{Maximum sequences per patient.}

The number of non-overlapping laboratory sequences sampled per patient was capped at ten for the general prediction task. Without this constraint, patients with prolonged treatment courses, who are also more likely to experience adverse outcomes, would contribute disproportionately to both training and evaluation, introducing survival bias and inflating performance estimates for endpoints correlated with treatment duration. A cap of ten sequences balances dataset size against this bias while preserving sufficient temporal coverage of each patient's disease course. For myelodysplastic syndromes, the sequence cap was increased to 20 per patient owing to low endpoint prevalence. At a cap of ten, positive sequences in the validation and test sets were insufficient for stable AUROC estimation. Increasing the cap to 20 yielded adequate positive case counts without materially altering the class balance of other diagnostic targets, as the additional sequences contributed primarily to patients with extended observation periods in whom therapy-related myeloid neoplasms are most clinically relevant. This design choice introduces additional survival bias for the MDS endpoint. Accordingly, MDS performance estimates should be interpreted with this enrichment in mind.

\subsubsection{Model architectures and training configuration}

Both the imputation and prediction models used a bidirectional transformer encoder based on the Qwen-3 architecture, trained from scratch without pretrained weights. Full architectural specifications and training hyperparameters are provided in Supplementary Tables~\ref{tab:imputation_architecture} and~\ref{tab:prediction_architecture}.

\subsubsection{Baseline models}

Baseline methodology is described in Methods. For each laboratory sequence, analyte measurements were converted into bag-of-words count vectors over the analyte vocabulary. Logistic regression was trained with logistic loss. XGBoost was trained as a gradient-boosted decision tree classifier. Both baselines were evaluated using the same metrics (AUROC and enrichment ratio) and bootstrap procedure as the transformer.

\subsubsection{Biomarker importance analysis}

The masking approach is described in Methods. The masking effect for analyte $a$ on diagnostic group $g$ was defined as:
\begin{equation}
\Delta\mathrm{AUROC}_{a,g} = \mathrm{AUROC}_{g}^{\text{masked}(a)} - \mathrm{AUROC}_{g}^{\text{full}},
\end{equation}
where $\mathrm{AUROC}_{g}^{\text{full}}$ denotes the area under the receiver operating characteristic curve with all analytes present and $\mathrm{AUROC}_{g}^{\text{masked}(a)}$ denotes the AUROC after masking analyte $a$. Negative values indicate that removing the analyte decreased predictive discrimination, with larger negative values reflecting greater importance.
 
Pairwise masking was performed analogously for all $\binom{40}{2} = 780$ analyte pairs. For each pair $(a_1, a_2)$ and each diagnostic group, both analytes were simultaneously replaced with zero across all timesteps, and the pairwise masking effect was computed as:
\begin{equation}
\Delta\mathrm{AUROC}_{(a_1,a_2),g} = \mathrm{AUROC}_{g}^{\text{masked}(a_1,a_2)} - \mathrm{AUROC}_{g}^{\text{full}}.
\end{equation}

\begin{table}[h]
\centering
\caption{\textbf{Imputation model architecture and training
configuration.}}
\label{tab:imputation_architecture}
\begin{tabular}{ll}
\hline
\textbf{Component} & \textbf{Specification} \\
\hline
Model backbone & Transformer encoder (Qwen-3, trained from scratch) \\
Hidden dimension & 256 \\
Number of layers & 4 \\
Number of attention heads & 2 \\
Number of Parameters & 4229160 \\
Token definition & One token per timestep encoding 39 analytes + age \\
Input projection & Linear projection of normalised analytes to hidden dim. \\
Prediction head & Linear layer mapping hidden dim. to analyte values \\
Loss function & Mean squared error on masked positions \\
Optimiser & Adam \\
Learning rate & 0.001 (cosine annealing schedule) \\
Batch size & 64 \\
Input normalisation & Per-analyte z-score normalisation (training set stats) \\
Missing value handling & Zero imputation with analyte-specific presence mask \\
Early stopping & MSE plateau ($\geq$40 epochs) \\
Typical training duration & $\sim$600 epochs \\
$P$ (mask individual analyte) & 0.2 \\
$P$ (mask entire timestep) & 0.05 \\
\hline
\end{tabular}
\end{table}

\begin{table}[h]
\centering
\caption{\textbf{Diagnosis prediction model architecture and
training configuration.}}
\label{tab:prediction_architecture}
\begin{tabular}{ll}
\hline
\textbf{Component} & \textbf{Specification} \\
\hline
Model backbone & Transformer encoder (Qwen-3, trained from scratch) \\
Hidden dimension & 64 \\
Number of layers & 4 \\
Number of attention heads & 2 \\
Number of Parameters & 484707 \\
Token definition & One token per timestep encoding 39 analytes + age + sex (single token per sequence) \\
Input projection & Linear projection of normalised analytes to hidden dimension \\
Token aggregation & Attention pooling across all token-wise last-layer transformer representations.\\
Prediction head & Linear layer applied to aggregated last-layer tokens \\
Loss function & focal loss, $\gamma$ 2.0 (class weights from training set) \\
Optimiser & Adam \\
Learning rate & 0.001 (cosine annealing schedule) \\
Batch size & 64 \\
Input normalisation & Per-analyte z-score normalisation (training set statistics) \\
Missing value handling & Pre-imputation via imputation model \\
Early stopping & AP plateau ($\geq$20 epochs) \\
Typical training duration & $\sim$40 epochs \\
Prediction horizon & 730 days \\
Cross-validation & 5 patient-level folds \\
Ensemble strategy & Mean logits across fold-specific models \\
\hline
\end{tabular}
\end{table}

\subsubsection{Evaluation metrics}

Model performance was evaluated using diagnosis-level classification metrics computed independently for each target diagnosis and subsequently aggregated across diagnostic groups.

\paragraph{Area under the receiver operating characteristic curve.}
Discrimination was assessed using the area under the receiver operating characteristic curve (AUROC), which quantifies the probability that a randomly selected positive instance receives a higher predicted score than a randomly selected negative instance. AUROC is threshold-independent and invariant to prevalence, which makes it comparable across endpoints but also insensitive to the precision achievable at any given prevalence. It is widely reported in the clinical prediction literature, facilitating comparison with existing models.

\paragraph{Average precision.}
To complement AUROC, we additionally computed average precision (AP) as implemented in \texttt{scikit-learn}. Whereas AUROC summarizes ranking performance over both classes and is invariant to prevalence, AP summarizes the precision--recall curve and therefore characterizes performance among the highest-ranked, positively-predicted patients, the operating region most relevant when positive cases are rare. AP is sensitive to prevalence: a model may attain a high AUROC while achieving low precision among its top-ranked predictions, a discrepancy that AUROC cannot reveal but AP makes explicit. Reporting both metrics thus separates a model's ability to rank cases (AUROC) from its ability to concentrate true cases among its most confident predictions (AP). AP is computed as a weighted sum of precision values at observed operating points, with weights given by the increase in recall between successive thresholds:
\begin{equation}
\mathrm{AP} = \sum_{k=1}^{K} \left( R_k - R_{k-1} \right) P_k,
\end{equation}
where $P_k$ and $R_k$ denote the precision and recall computed at the $k$-th distinct decision threshold after sorting predictions in descending order, and $R_0 = 0$. This formulation computes a non-interpolated approximation of the area under the precision--recall curve using a step-function summation, which avoids the overestimation associated with linear (trapezoidal) interpolation. Average precision provides a threshold-independent performance summary and is appropriate in settings with substantial class imbalance.

The enrichment ratio, defined as AP divided by endpoint prevalence, quantifies the gain in top-ranked precision relative to a random classifier, whose expected AP equals the prevalence. Values above one therefore indicate that true cases are concentrated among the highest-ranked patients more than chance would predict, which is informative under severe class imbalance, where a moderate AUROC can still correspond to substantial enrichment.

\paragraph{Diagnostic group-level evaluation.}

For predefined diagnostic groups comprising multiple related diagnoses, group-level prediction scores were constructed by taking the maximum predicted probability across all constituent diagnoses for each sample:
\begin{equation}
\hat{y}_g = \max_{d \in g} \hat{y}_d.
\end{equation}
Group-level ground-truth labels were defined analogously:
\begin{equation}
y_g = \max_{d \in g} y_d.
\end{equation}
AUROC and average precision were then computed for each diagnostic group using these derived predictions and labels.

\paragraph{Confidence intervals.}

Uncertainty in performance estimates was quantified using nonparametric bootstrap resampling of the held-out test set. The test set was resampled with replacement at the sequence level 1000 times, and AUROC and average precision were recomputed for each diagnosis and diagnostic group on each replicate. Performance was summarised as the median across replicates, and confidence intervals were obtained from the empirical bootstrap distribution using percentile-based intervals.

Unless stated otherwise, all metrics were computed on the held-out test set at the diagnosis level and reported either individually or as macro-averaged values across diagnoses.

\subsubsection{Dataset summary}

Dataset splits and cohort sizes are reported in Methods. Per-endpoint prevalences for all 162 individual ICD codes and diagnostic groups are provided in Supplementary Tables~\ref{tab:performance_comparison} and~\ref{tab:performance_comparison_grouped}.

\end{document}